\documentclass{article}

\PassOptionsToPackage{numbers, compress}{natbib}

\usepackage[final]{neurips_2022}

\usepackage{marvosym}
\usepackage[utf8]{inputenc} 
\usepackage[T1]{fontenc}    
\usepackage{url}            
\usepackage{booktabs}       
\usepackage{amsfonts}       
\usepackage{nicefrac}       
\usepackage{microtype}      
\usepackage{xcolor}         
\usepackage{graphicx}
\usepackage{mathrsfs}
\makeatletter
  \newcommand\figcaption{\def\@captype{figure}\caption}
  \newcommand\tabcaption{\def\@captype{table}\caption}
\makeatother
\usepackage{makecell}

\usepackage{times}
\usepackage{epsfig}
\usepackage{amsmath}
\usepackage{amssymb}
\usepackage{wrapfig}
\usepackage{amsthm}
\usepackage{multirow}
\usepackage{wrapfig}
\usepackage{tabu}
\usepackage{bbm}
\definecolor{citecolor}{HTML}{2980b9}
\definecolor{linkcolor}{HTML}{c0392b}
\usepackage[pagebackref=true,breaklinks=true,colorlinks,bookmarks=false,citecolor=citecolor,linkcolor=linkcolor]{hyperref}
\usepackage{xcolor}         
\usepackage{color, colortbl}
\usepackage{float}
\usepackage{adjustbox}
\usepackage{chngcntr}
\setcitestyle{square,numbers}

\title{Point-M2AE: Multi-scale Masked Autoencoders\\ for Hierarchical Point Cloud Pre-training}

\author{Renrui Zhang$^{1,2}$, Ziyu Guo$^{2}$, Rongyao Fang$^{1}$,\vspace{0.1cm}\\ \textbf{Bin Zhao$^2$, Dong Wang$^2$, Yu Qiao$^2$, Hongsheng Li$^{1,3}$, Peng Gao\textsuperscript{\Letter}$^{2}$}\vspace{0.3cm}\\
  $^1$ CUHK-SenseTime Joint Laboratory, The Chinese University of Hong Kong,\\
  $^2$ Shanghai AI Laboratory,
  $^3$ Centre for Perceptual and Interactive Intelligence Limited\vspace{0.1cm}\\
  \texttt{\{zhangrenrui, gaopeng\}@pjlab.org.cn} \\
  \texttt{hsli@ee.cuhk.edu.hk} \\
}

\begin{document}

\maketitle

\begin{abstract}
Masked Autoencoders (MAE) have shown great potentials in self-supervised pre-training for language and 2D image transformers. However, it still remains an open question on how to exploit masked autoencoding for learning 3D representations of irregular point clouds. In this paper, we propose \textbf{Point-M2AE}, a strong \textbf{M}ulti-scale \textbf{M}AE pre-training framework for hierarchical self-supervised learning of 3D point clouds. Unlike the standard transformer in MAE, we modify the encoder and decoder into pyramid architectures to progressively model spatial geometries and capture both fine-grained and high-level semantics of 3D shapes. For the encoder that downsamples point tokens by stages, we design a multi-scale masking strategy to generate consistent visible regions across scales, and adopt a local spatial self-attention mechanism during fine-tuning to focus on neighboring patterns. By multi-scale token propagation, the lightweight decoder gradually upsamples point tokens with complementary skip connections from the encoder, which further promotes the reconstruction from a global-to-local perspective. 
Extensive experiments demonstrate the \textit{state-of-the-art} performance of Point-M2AE for 3D representation learning. With a frozen encoder after pre-training, Point-M2AE achieves \textbf{92.9\%} accuracy for linear SVM on ModelNet40, even surpassing some fully trained methods. By fine-tuning on downstream tasks, Point-M2AE achieves \textbf{86.43\%} accuracy on ScanObjectNN, \textbf{+3.36\%} to the second-best, and largely benefits the few-shot classification, part segmentation and 3D object detection with the hierarchical pre-training scheme. Code is available at \url{https://github.com/ZrrSkywalker/Point-M2AE}.

\end{abstract}

\section{Introduction}

Learning to represent from unlabeled data without annotations, known as self-supervised learning, has attained great success in natural language processing~\cite{bert,gpt1,gpt2,gpt3}, computer vision~\cite{he2020momentum,chen2020simple,chen2021exploring,mae} and multi-modality learning~\cite{clip,zhang2021pointclip,align}. By pre-training on the large-scale raw data, the networks are endowed with robust representation abilities and can significantly benefit downstream tasks with fine-tuning. Motivated by masked language modeling~\cite{gpt1,bert}, MAE~\cite{mae} and some other methods~\cite{xie2021simmim,zhou2021ibot,baevski2022data2vec} adopt asymmetric encoder-decoder transformers~\cite{vit} to apply masked autoencoding for self-supervised learning on 2D images. They represent the input image as multiple local patches, and randomly mask them with a high ratio to build the pretext task for reconstruction. 
Specifically, the encoder aims at capturing high-level latent representations from limited visible patches, and the lightweight decoder is forced to reconstruct the RGB values of masked patches on top. 
Despite its superiority on grid-based 2D images, we ask the question: can MAE-style masked autoencoding be adapted to irregular point clouds as a powerful 3D representation learner?
\begin{figure*}[t!]
  \centering
    \includegraphics[width=1\textwidth]{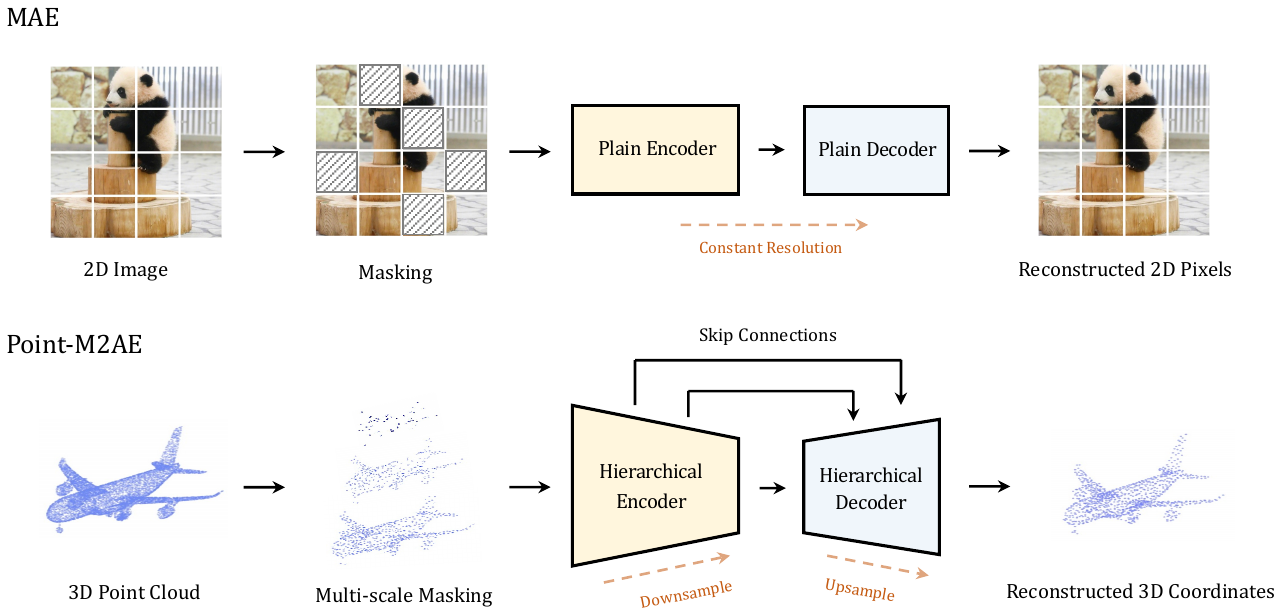}
    \vspace{0.01cm}
   \caption{\textbf{Comparison of MAE (Top) and our Point-M2AE (Bottom).} MAE for 2D image pre-training adopts standard transformer of the plain encoder and decoder, while Point-M2AE introduces a hierarchical transformer with skip connections for multi-scale point cloud pre-training.}
    \label{fig1}
\end{figure*}

To tackle this challenge, we propose \textbf{M}ulti-scale \textbf{M}asked autoencoders for learning the hierarchical representations of point clouds via self-supervised pre-training, termed as Point-M2AE. We represent a point cloud as a set of point tokens depicting different spatial local regions, and inherit MAE's pipeline to first encode visible point tokens and then reconstruct the masked 3D coordinates. 
Different from 2D images, masked autoencoding for 3D point clouds has three characteristics to be specially considered. 
\textit{\textbf{Firstly,}} it is critical to understand the relations between local parts and the overall 3D shapes, which have strong geometric and semantic dependence. As examples, the network can recognize an airplane starting from its wing, or segment the wing's part from the airplane's global feature. Therefore, we regard the standard transformer with the plain encoder and decoder is sub-optimal for capturing such local-global spatial relations in 3D, which directly downsamples the input into a low-resolution representation as shown in Figure~\ref{fig1} (Top). We modify both the encoder and decoder into multi-stage hierarchies for progressively encoding multi-scale features of point clouds, constructing an asymmetric U-Net~\cite{unet} like architecture in Figure~\ref{fig1} (Bottom). 
\textit{\textbf{Secondly,}} as our Point-M2AE encodes multi-scale point clouds unlike the single-scale 2D images, the unmasked visible regions are required to be both block-wise within one scale and consistent across scales, which are respectively for reserving complete local geometries and ensuring coherent feature learning for the network. For this, we introduce a multi-scale masking strategy, which generates random masks at the final scale with a high ratio (e.g., 80\%), and back-projects the unmasked positions to all preceding scales. \textit{\textbf{Thirdly,}} to better reconstruct 3D geometries from a local-to-global perspective, we utilize skip connections to complement the decoder with fine-grained information from the corresponding stages of the encoder.
During fine-tuning on downstream tasks, we also adopt a local spatial self-attention mechanism with increasing attention scopes for point tokens at different stages of the encoder, which refocus each token within neighboring detailed structures.

By the multi-scale pre-training, Point-M2AE can encode point clouds from local-to-global hierarchies and then reconstructs the masked coordinates from global-to-local perspectives, which learns powerful 3D representations and performs superior transfer ability. After self-supervised pre-training on ShapeNet~\cite{chang2015shapenet}, Point-M2AE achieves 92.9\% classification accuracy for linear SVM on ModelNet40~\cite{modelnet40} with the frozen encoder, which surpasses the runner-up CrossPoint~\cite{afham2022crosspoint} by +1.2\% and even outperforms some fully supervised methods. By fine-tuning on various downstream tasks, Point-M2AE achieves 86.43\% (+3.36\%) accuracy on ScanObjectNN~\cite{scanobjectnn} and 94.0\% (+0.8\%) accuracy on ModelNet40~\cite{modelnet40} for shape classification, 86.51\% (+0.91\%) instance mIoU on ShapeNetPart~\cite{shapenetpart} for part segmentation, and 95.0\% (+2.7\%) accuracy on 10-way 20-shot ModelNet40 for few-shot classification. Our multi-scale masked autoencoding also benefits the 3D object detection on ScanNetV2~\cite{ScanNetV2} by +1.3\% AP$_{25}$ and +1.3\% AP$_{50}$, which provides the detection backbone with a hierarchical understanding of the point clouds. 

We summarize the contributions of our paper as follows:
\begin{enumerate}
    \item We propose Point-M2AE, a strong masked autoencoding framework, which conducts hierarchical point cloud encoding and reconstruction for better learning multi-scale spatial geometries of 3D shapes.
    
    \item We introduce a U-Net like transformer architecture for MAE-style pre-training on point clouds, and adopt a multi-scale masking strategy to generate consistent visible regions across scales.
    
    \item Point-M2AE achieves \textit{state-of-the-art} performance for transfer learning on various downstream tasks, which indicates our approach to be a powerful representation learner for 3D point clouds.
\end{enumerate}

\section{Related Work}
\paragraph{Pre-training by Masked Modeling.}
Compared to contrastive learning methods~\cite{he2020momentum,chen2020simple,chen2021exploring} that learn from inter-sample relations, self-supervised pre-training by masked autoencoding builds the pretext tasks to predict the masked parts of the input signals. The series of GPT~\cite{gpt1,gpt2,gpt3} and BERT~\cite{devlin2018bert} apply masked modeling to natural language processing and achieve extraordinary performance boost on downstream tasks with fine-tuning. Inspired by this, BEiT~\cite{bao2021beit} proposes to match image patches with discrete tokens via dVAE~\cite{dvae} and pre-train a standard vision transformer~\cite{vit,pointbert} by masked image modeling. On top of that, MAE~\cite{mae} directly reconstructs the raw pixel values of masked tokens and performs great efficiency with a high mask ratio. The follow-up works further improve the performance of MAE by momentum encoder~\cite{zhou2021ibot}, contrastive learning~\cite{baevski2022data2vec}, and modified reconstruction targets~\cite{wei2021masked}. For self-supervised pre-training on 3D point clouds, the masked autoencoding has not been widely adopted. Similar to BEiT, Point-BERT~\cite{pointbert} utilizes dVAE to map 3D patches to tokens for masked point modeling, but heavily relies on constrastive learning~\cite{he2020momentum}, complicated data augmentation, and the costly two-stage pre-training. In contrast, our Point-M2AE is a pure masked autoencoding method of one-stage pre-training, and follows MAE to reconstruct the input signals without dVAE mapping. Different from previous MAE methods adopting standard plain transformer, we propose a hierarchical transformer architecture along with the multi-scale masking strategy to better learn a strong and generic representation for 3D point clouds.

\paragraph{Self-supervised Learning for Point Clouds.}
3D representation learning without annotations has been widely studied in recent years. Mainstream methods mainly build the pretext tasks to reconstruct the transformed input point cloud based on the encoded latent vectors, such as rotation~\cite{poursaeed2020self}, deformation~\cite{achituve2021self}, rearranged parts~\cite{sauder2019self} and occlusion~\cite{occo}. From another perspective, PointContrast~\cite{pointcontrast} utilizes contrastive learning between features of the same points from different views to learn discriminative 3D representations. DepthContrast~\cite{depthcontrast} further extends the contrast for depth maps of different augmentations. CrossPoint~\cite{afham2022crosspoint} conducts cross-modality contrastive learning between point clouds and their corresponding rendering images to acquire rich self-supervised signals. Point-BERT~\cite{pointbert} and Point-MAE~\cite{pang2022masked} respectively introduce BERT-style~\cite{bert} and MAE-style~\cite{mae} pre-training schemes for 3D point clouds with standard transformer networks and performs competitively on various downstream tasks, but both of them can only encode point clouds with a single resolution and ignores the local-global relations between 3D shapes. In this paper, we propose Point-M2AE, an MAE-style framework with a hierarchical transformer for multi-scale point cloud pre-training. We achieve \textit{state-of-the-art} downstream performance by learning the multi-scale representation of point clouds.

\section{Method}
\begin{figure*}[t!]
  \centering
    \includegraphics[width=1\textwidth]{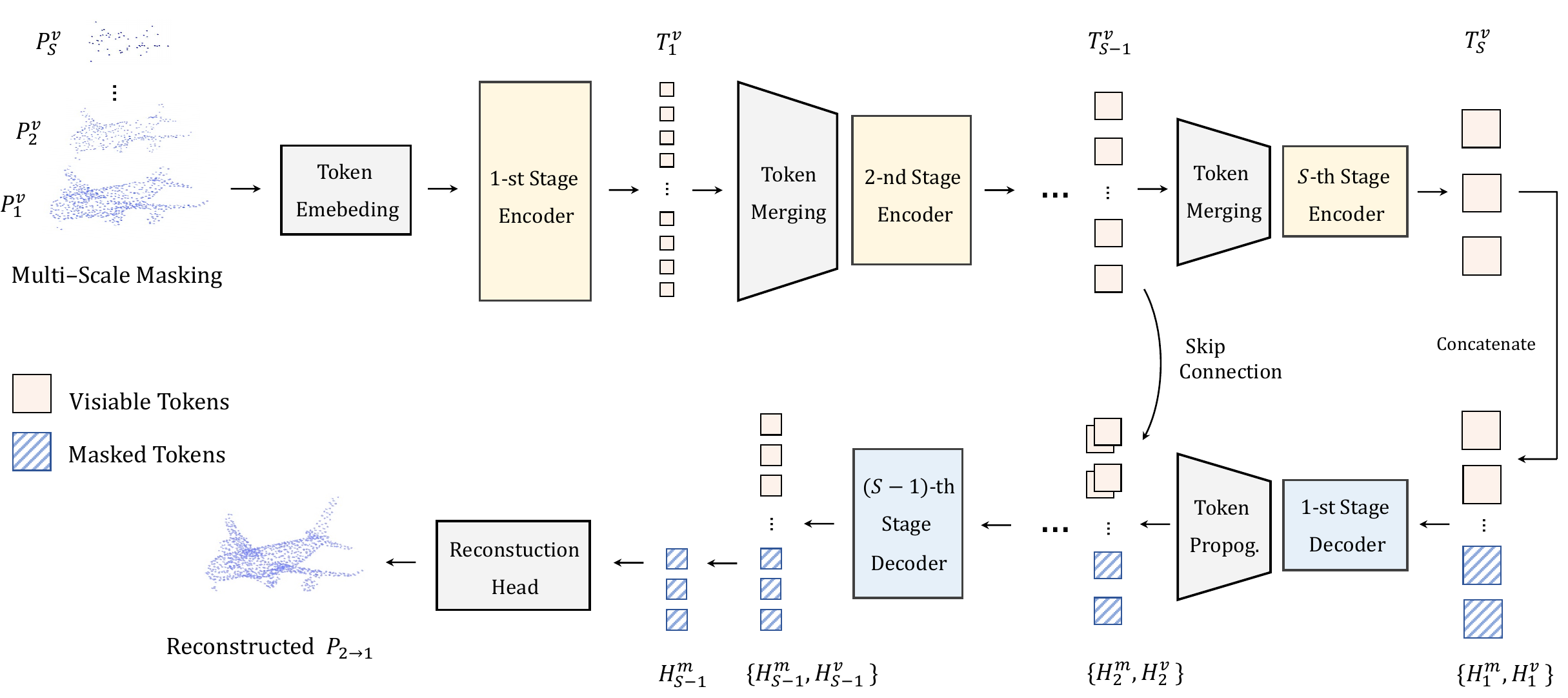}
    \vspace{0.01cm}
   \caption{\textbf{Overall pipeline of Point-M2AE.} After the multi-scale masking, we embed point tokens at the $1$-st scale and feed the visible ones into a hierarchical encoder-decoder transformer, which captures both high-level semantics and fine-grained patterns of the point cloud during pre-training.}
    \label{fig2}
    \vspace{-0.3cm}
\end{figure*}

The overall pipeline of Point-M2AE is shown in Figure~\ref{fig2}, where we encode and reconstruct the point cloud by a hierarchical network architecture.
In Section~\ref{mmask}, We first introduce the masking strategy of Point-M2AE with multi-scale representations of point clouds. Then in Section~\ref{hencoder} and Section~\ref{hdecoder}, we present the details of our encoder and decoder with multi-stage hierarchies.

\subsection{Multi-scale Masking}
\label{mmask}
\begin{figure*}[t!]
  \centering
    \includegraphics[width=0.95\textwidth]{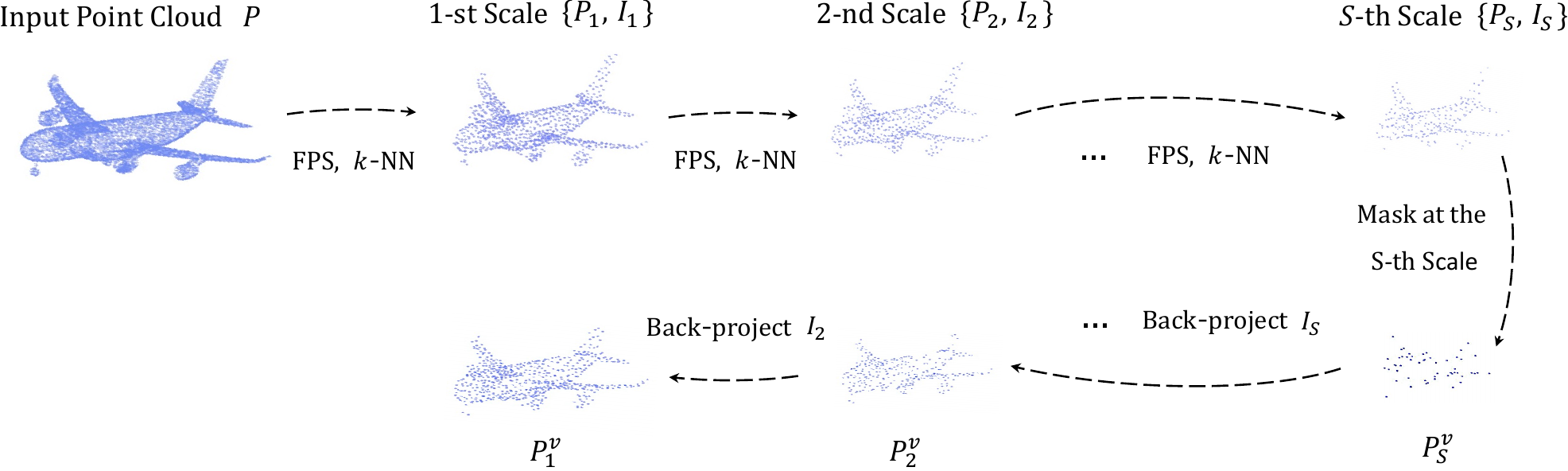}
    \vspace{0.1cm}
   \caption{\textbf{Multi-scale masking strategy.} To obtain a consistent visible regions across scales, we first represent the input point cloud by multi-scale coordinates and generate the random mask at the highest one. Then, we back-project the unmasked visible positions to all earlier scales.}
    \label{fig3}
\end{figure*}

To build a U-Net~\cite{unet} like masked autoencoder for hierarchical learning, we encode the point cloud by $S$ scales with different number of points at each scale, and correspondingly modify the standard plain encoder into the $S$-stage architecture. Following MAE, we embed the point cloud into discrete point tokens and randomly mask them for reconstruction. Importantly, for irregular-distributed points in the multi-scale architecture, the unmasked visible spatial regions are required to be consistent not only within one scale, but also across different scales. This is because the block-wise parts of 3D shapes tend to preserve more complete fine-grained geometries, and the unmasked positions are better to be shared across all scales for coherent feature learning of the encoder.
Therefore, as shown in Figure~\ref{fig3}, we first construct the $S$-scale coordinate representations of the input point cloud and back-project the random masks from the final $S$-th scale to the earlier scales to avoid fragmented visible parts.


\paragraph{$S$-scale Representations.}
We denote the input point cloud as $P \in \mathbb{R}^{N\times 3}$ and regard it as the $0$-th scale. For the $i$-th scale, $1 \leq i \leq S$, we utilize Furthest Point Sampling (FPS) to downsample the points from the $(i-1)$-th scale, which produces seed points $P_i \in \mathbb{R}^{N_i\times 3}$ for scale $i$ of $N_i$ points. Then, we adopt $k$ Nearest-Neighbour ($k$-NN) to aggregate the neighboring $k$ points for each seed point and obtain the neighbor indices $I_i \in \mathbb{R}^{N_i\times k}$. By successively downsampling and grouping, we acquire the $S$-scale representations $\{P_i, I_i \}_{i=1}^S$ of the input point cloud, where the number of points $N_i$ gradually decreases and the inclusion relations between scales are recorded in $I_i$. 

\paragraph{Back-projecting Visible Positions.}
For seed points $P_S$ at the final $S$-th scale, we randomly mask them with a large proportion (e.g., 80\%) and denote the remaining visible points as $P^v_S\in \mathbb{R}^{N_S^v\times 3}$ of $N_S$ points. We then back-project the unmasked positions $P^v_S$ to ensure the consistent visible regions across scales.
For the $i$-th scale, $1 \leq i < S$, we retrieve all the $k$ nearest neighbors of $P^v_{i+1}$ from the indices $I_{i+1}$ to serve as the visible positions $P^v_{i}$, and mask the others. By recursively back-projecting, we obtain the visible and masked positions of all $S$ scales, denoted as $\{P^v_i, P^m_i\}_{i=1}^S$, where $P^v_i\in \mathbb{R}^{N_{i}^v\times 3}$, $P^m_i\in \mathbb{R}^{N_{i}^m\times 3}$ and $N_i = N_{i}^v + N_{i}^m$.

\subsection{Hierarchical Encoder}
\label{hencoder}
Based on the multi-scale masking, we embed the initial tokens of visible points $P_1^v$ for the $1$-st scale and them into the hierarchical encoder with $S$ stages. Every stage is equipped with $K$ stacked encoder blocks, and each block contains a self-attention layer and a Feed Forward Network (FFN) of MLP layers. Between every two consecutive stages, we introduce spatial token merging modules to aggregate adjacent visible tokens and enlarge receptive fields for downsampling the point clouds.

\paragraph{Token Embedding and Merging.}
Indexed by $I_1$, we utilize a mini-PointNet~\cite{qi2017pointnet} to extract and fuse the features of every seed point from $P_1^v \in \mathbb{R}^{N_1^v\times 3}$ with its $k$ nearest neighbors. After that, we obtain the initial point tokens $T_1^v \in \mathbb{R}^{N_1^v\times C_1}$ for the 1-st stage of the encoder, which embeds $N_1^e$ local patterns of the 3D shape.
Between the $(i-1)$-th and $i$-th stages, $1 < i \leq S$, we merge $T_{i-1}^v \in \mathbb{R}^{N_{i-1}\times C_{i-1}}$ to acquire the downsampled point tokens for the $i$-th stage. We utilize MLP layers and a max pooling to integrate every $k$ tokens nearest to $P_i^v$ indexed by $I_i$, which outputs $T_i^v \in \mathbb{R}^{N_i\times C_i}$. 
Due to our multi-scale masking, the merged $T_i^v$ corresponds to the same visible parts of $T_{i-1}^v$, which enables the consistent feature encoding across different scales.
For larger $i$ of deeper stages, we set higher feature dimension $C_i$ to encode spatial geometries with richer semantics. 
\paragraph{Local Spatial Self-Attention.}
During pre-training, we expect point tokens in the multi-stage encoder to capture global cues for 3D shapes, which benefits the reconstruction of masked parts.
However, when fine-tuning on downstream tasks without masked autoencoding, point tokens in the shallower stages are better to mainly focus on local information and not to be disturbed by long-range signals, referring to the inductive bias of 3D locality~\cite{qi2017pointnet++}. Thus, during fine-tuning, we modify the original self-attention layer in the encoder with a local spatial constraint that only neighboring tokens within a ball query would be available for attention calculation. As the point tokens are downsampled by stages, we set increasing radii $\{r_i\}_{i=1}^S$ of multi-scale ball queries for gradually expanding the attention scopes, which fulfills the local-to-global feature aggregation scheme.
\subsection{Hierarchical Decoder}
\label{hdecoder}
Via the hierarchical encoder, we obtain the encoded visible tokens $\{T^v_i\}_{i=1}^S$ of all scales. Starting from the highest $S$-th scale, we assign a shared learnable mask token to all the masked positions $P_S^m$, and concatenate them with the visible tokens $T^v_S$.
We denote them as $\{H_1^v, H_1^m\}$ with coordinates $\{P_S^v, P_S^m\}$, which serve as the input of the hierarchical decoder.
We design the decoder to be lightweight with $S-1$ stages and only one decoder block for each stage, which enforces the encoder to embed more semantics of the point clouds. Each decoder block consists of a vanilla self-attention layer and an FFN. We do not apply the local constraint to the attention in the decoder, since a global understanding between visible and mask tokens is crucial to the reconstruction.
\paragraph{Point Token Upsampling.}
We upsample the point tokens between stages to progressively recover the fine-grained geometries of 3D shapes before reconstruction. We regulate that the $j$-th stage of the decoder corresponds to the $(S+1-j)$-th stage of the encoder, both of which contain point tokens of the same $(S+1-j)$-th scale with the feature dimension $C_{S+1-j}$.
Between the $(j-1)$-th and $j$-th stage, $1 < j \leq S-1$, we upsample the tokens $\{H_{j-1}^v, H_{j-1}^m\}$ from the coordinates $\{P_{S+2-j}^v, P_{S+2-j}^m\}$ into $\{P_{S+1-j}^v, P_{S+1-j}^m\}$ via the token propagation module. Specifically, we obtain the $k$ nearest neighbors of each point token in $\{H_{j-1}^v, H_{j-1}^m\}$ indexed by $I_{S+2-j}$, and recover their neighbors' features by weighted interpolation referring to PointNet++~\cite{qi2017pointnet++}, which generates the tokens $\{H_{j}^v, H_{j}^m\}$ of the $j$-th stage.

\paragraph{Skip Connections.} 
To further complement the fine-grained geometries, we channel-wisely concatenate the visible tokens $H^v_j \in \mathbb{R}^{N_{S+1-j}\times C_{S+1-j}}$ of the decoder with $T^v_{S+1-j} \in \mathbb{R}^{N_{S+1-j}\times C_{S+1-j}}$ from the corresponding $(S+1-j)$-th stage of the encoder via skip connections, and adopt a linear projection layer to fuse their features. For the mask tokens $H^m_j$, we keep them unchanged, since the encoder only contains visible tokens without the masked ones.

\paragraph{Point Reconstruction.}
After $S-1$ stages of the decoder, we acquire $\{H^v_{S-1}, H^m_{S-1}\}$ with coordinates $\{P^v_{2}, P^m_{2}\}$ and reconstruct the masked values from the mask tokens $H^m_{S-1}$. Other than predicting values at the $0$-th scale of the input point cloud $P$, we reconstruct the coordinates of $P^m_1$, namely, recovering the masked positions of the $1$-st scale $P^m_{1} \in \mathbb{R}^{N^m_{1}\times 3}$ from the $2$-nd scale $P^m_{2} \in \mathbb{R}^{N_{2}^m\times 3}$. This is because $\{P^v_{1}, P^m_{1}\}$ of the 1-st scale could well represent the overall 3D shape and simultaneously preserve enough local patterns, which already constructs a comparatively challenging pretext task for pre-training. If we further upsample $\{H^v_{S-1}, H^m_{S-1}\}$ into $\{H^v_{S}, H^m_{S}\}$ and reconstruct the masked raw points from $P^m_{1}$, the extra spatial noises and computational overhead would adversely influence our performance and efficiency. Therefore, for every token in $H^m_{S-1} \in \mathbb{R}^{N_{2}^m\times C_2}$, we reconstruct its $k$ nearest neighbors recorded in $I_2$ by a reconstruction head of one linear projection layer and compute the loss by $l_2$ Chamfer Distance~\cite{chamfer},
formulated as,
\begin{align}
    &\widehat{P}_{2\rightarrow 1}^m = \mathrm{Linear}(H_{S-1}^m), \ \ \ \text{where}\ \widehat{P}_{2\rightarrow 1}^m \in \mathbb{R}^{N_2^m\times k \times 3}, \\
    &\mathcal{L}_{CD} = \text{ChamferDistance}({P}_{2\rightarrow 1}^m, \widehat{P}_{2\rightarrow 1}^m),
\end{align}
where $\widehat{P}_{2\rightarrow 1}^m$ and $P_{2\rightarrow 1}^m$ denote the predicted and ground-truth reconstruction coordinates from the $2$-nd scale to the $1$-st scale. We only utilize $\mathcal{L}_{CD}$ for supervision without contrastive loss to conduct a pure masked autoencoding for self-supervised pre-training.

\section{Experiments}

\begin{figure*}
\begin{minipage}[t!]{0.42\linewidth}
\centering
\small
\vspace{-0.35cm}
\tabcaption{\textbf{Linear evaluation on ModelNet40~\cite{modelnet40} by SVM.} We report different self-supervised learning methods and underline the second-best one.}
\vspace{0.4cm}
\label{svm}
\begin{adjustbox}{width=\linewidth}
	\begin{tabular}{lc}
	\toprule
		Method &Acc. (\%)\\
        \cmidrule(lr){1-1} \cmidrule(lr){2-2}
        3D-GAN~\cite{3dgan}  &83.3  \\
        Latent-GAN~\cite{latentgan}   & 85.7  \\
        SO-Net~\cite{sonet}   &87.3  \\
        FoldingNet~\cite{foldingnet}  &88.4 \\
        MAP-VAE~\cite{mapvae}  & 88.4 \\
		VIP-GAN~\cite{vipgan}   &90.2 \\
		\cmidrule(lr){1-2}
		DGCNN + Jiasaw~\cite{jiasaw}  &90.6  \\
		DGCNN + OcCo~\cite{occo}  &90.7  \\
        DGCNN + CrossPoint~\cite{afham2022crosspoint}  &\underline{91.2} \\
        \cmidrule(lr){1-2}
        Transformer + OcCo~\cite{pointbert} &89.6 \\
        Point-BERT~\cite{pointbert}  &87.4\vspace{0.05cm}\\
		\rowcolor{gray!8}\textbf{Point-M2AE}  &\textbf{ 92.9} \vspace{0.1cm}\\
		\textit{Improvement}  &\textcolor{blue}{+1.7} \\
	\bottomrule
	\end{tabular}
\vspace*{2pt}
\end{adjustbox}
\end{minipage}\quad
\begin{minipage}[t!]{0.56\linewidth}
\centering
\small
\vspace{-0.35cm}
\tabcaption{\textbf{Shape classification on ModelNet40~\cite{modelnet40}.} `\#points' and `Acc.' denote the number of points for training and the overall accuracy. {[S]} represents fine-tuning after self-supervised pre-training.}
\vspace{0.3cm}
\label{modelnet_cls}
\begin{adjustbox}{width=\linewidth}
\centering
	\begin{tabular}{lcc}
	\toprule
		Method &\#points  &Acc. (\%)\\
		\cmidrule(lr){1-1} \cmidrule(lr){2-2} \cmidrule(lr){3-3} 
         PointNet~\cite{qi2017pointnet}  &1k &89.2 \\ 
        PointNet++~\cite{qi2017pointnet++}  &1k &90.5 \\
        PointCNN~\cite{li2018pointcnn}  &1k &92.2 \\
        {[S]} SO-Net~\cite{sonet}  &5k &92.5 \\
        DGCNN~\cite{dgcnn}  &1k &92.9 \\
        PCT~\cite{guo2021pct}  &1k &93.2 \\
        Point Transformer~\cite{pointtransformer}  &- &93.7 \\
        \cmidrule(lr){1-3}
        Transformer~\cite{pointbert}  &1k &91.4\\
        {[S] Transformer + OcCo}~\cite{pointbert}  &1k &92.1\vspace{0.05cm}\\
        {[S] Point-BERT}~\cite{pointbert}  &1k &93.2\\
        {[S] Point-BERT}  &4k &93.4\\
        {[S] Point-BERT}  &8k &93.8\vspace{0.1cm}\\
        \rowcolor{gray!8}\textbf{[S] Point-M2AE}  &\textbf{1k} &\textbf{94.0}\vspace{0.1cm}\\
	\bottomrule
	\end{tabular}
\end{adjustbox}
\end{minipage}
\end{figure*}

In Section~\ref{pretrain} and Section~\ref{downstream}, we introduce the pre-training experiments of Point-M2AE and report the fine-tuning performance on various downstream tasks. We also conduct ablation studies in Section~\ref{ablation} to validate the effectiveness of our approach.

\subsection{Self-supervised Pre-training}
\label{pretrain}
\paragraph{Settings.}
We pre-train our Point-M2AE on ShapeNet~\cite{chang2015shapenet} dataset, which contains 57,448 synthetic 3D shapes of 55 categories. 
We set the stage number $S$ as 3, and construct a 3-stage encoder and a 2-stage decoder for hierarchical learning. We adopt 5 blocks in each encoder stage, but only 1 block per stage for the lightweight decoder. For the 3-scale point clouds, we set the point numbers and token dimensions respectively as \{512, 256, 64\} and \{96, 192, 384\}. We also set different $k$ for the $k$-NN at different scales, which are \{16, 8, 8\}. We mask the highest scale of point clouds with a high ratio of 80\% and set 6 heads for all the attention modules. The detailed training settings are in Appendix.

\vspace{-0.1cm}
\paragraph{Linear SVM.}
After pre-training on ShapeNet, we test the 3D representation capability of Point-M2AE via linear evaluation on ModelNet40~\cite{modelnet40}. We sample 1,024 points from each 3D shape of ModelNet40 and utilize our frozen encoder to extract their features. On top of that, we train a linear SVM and report the classification accuracy in Table~\ref{svm}. As shown, Point-M2AE achieves the best performance among all existing self-supervised methods for point clouds, and surpasses the second-best CrossPoint~\cite{afham2022crosspoint} by +1.7\%. Point-M2AE also exceeds Point-BERT~\cite{pointbert} by +5.5\%, which is a masked point modeling method with a MoCo loss~\cite{he2020momentum} but adopts a standard transformer and conducts single-scale learning. It is worth noting that even if we freeze all our parameters, Point-M2AE with 92.9\% accuracy still outperforms many fully trained methods on ModelNet40, e.g., 90.5\% by PointNet++~\cite{qi2017pointnet++}, 92.8\% by DensePoint~\cite{densepoint}, etc. The experiments fully demonstrate the superior 3D representation capacity of our Point-M2AE.

\subsection{Downstream Tasks}
\label{downstream}
For fine-tuning on downstream tasks, we discard the hierarchical decoder in pre-training and append different heads onto the hierarchical encoder for different tasks.

\paragraph{Shape Classification.}
We fine-tune Point-M2AE on two shape classification datasets: the widely adopted ModelNet40~\cite{modelnet40} and the challenging ScanObjectNN~\cite{scanobjectnn}.
For local spatial attention layers, we set the ball queries' radii of 3-scale point clouds as \{0.32, 0.64, 1.28\}.
We follow Point-BERT to use the voting strategy~\cite{rscnn} for fair comparison on ModelNet40. To handle the noisy spatial structures, we increase $k$ of $k$-NN into \{32, 16, 16\} for ScanObjectNN to encode local patterns with larger receptive fields. 
As reported in Table~\ref{modelnet_cls}, Point-M2AE achieves 94.0\% accuracy on ModelNet40 with 1024 points per sample, which surpasses Point-BERT fine-tuned with 1024 points by +0.8\% and 8192 points by +0.2\%. For ScanObjectNN in Table~\ref{scan_cls}, our Point-M2AE outperforms the second-best Point-BERT by a significant margin, +3.79\%, +0.69\% and +3.36\%, respectively for the three splits, indicating our great advantages under complex circumstances by multi-scale encoding. As ScanObjectNN of real-world scenes has a large semantic gap with the pre-trained synthetic ShapeNet, Point-M2AE also exerts strong transfer ability to understand point clouds of another domain.

\begin{table}[t!]
\vspace{0.2cm}
\small
\centering
\tabcaption{\textbf{Shape classification on ScanObjectNN~\cite{scanobjectnn}}. We report the accuracy (\%) on the three splits of ScanObjectNN. {[S]} represents fine-tuning after self-supervised pre-training.}
\label{scan_cls}
\begin{adjustbox}{width=0.74\linewidth}
	\begin{tabular}{lccc}
	\toprule
		\makecell*[c]{Method} &OBJ-BG &OBJ-ONLY &PB-T50-RS\\
		\cmidrule(lr){1-1} \cmidrule(lr){2-2} \cmidrule(lr){3-3} \cmidrule(lr){4-4}
	    PointNet~\cite{qi2017pointnet}  &73.3 &79.2 &68.0\\
	    PointNet++~\cite{qi2017pointnet++} &82.3 &84.3 &77.9\\
	    DGCNN~\cite{dgcnn} &82.8 &86.2 &78.1\\
	    PointCNN~\cite{li2018pointcnn} &86.1 &85.5 &78.5\\
	   \cmidrule(lr){1-4}
	    Transformer~\cite{pointbert} &79.86 &80.55 &77.24  \\
	    {[S]} Transformer + OcCo~\cite{pointbert} &84.85 &85.54 &78.79\vspace{0.05cm}\\
	    {[S]} Point-BERT~\cite{pointbert} &87.43 &88.12 &83.07\vspace{0.1cm}\\
	    \rowcolor{gray!8} \textbf{[S] Point-M2AE} &\textbf{91.22} &\textbf{88.81}  &\textbf{86.43}\vspace{0.1cm}\\
	     \textit{Improvement} &\textcolor{blue}{+3.79} &\textcolor{blue}{+0.69} &\textcolor{blue}{+3.36} \\
	  \bottomrule
	\end{tabular}
\end{adjustbox}
\end{table}

\paragraph{Part Segmentation.}
We evaluate Point-M2AE for part segmentation on ShapeNetPart~\cite{shapenetpart}, which predicts per-point part labels and requires detailed understanding for local patterns. We adopt an extremely simple segmentation head to validate the effectiveness of our pre-training for well capturing both high-level semantics and fine-grained details. By the hierarchical encoder, we obtain $3$-scale point tokens of \{512, 256, 64\} points, and perform feature propagation in PointNet++~\cite{qi2017pointnet++} to independently upsample the tokens into 2048 points of the input point cloud. Then, we concatenate the upsampled $3$-scale features for each point and predict the part label by stacked linear projection layers. As reported in Table~\ref{seg}, Point-M2AE achieves the best 86.51\% instance mIoU with the simple segmentation head, surpassing the second-best Point-BERT by +0.91\%. Note that Point-BERT~\cite{pointbert} and other methods~\cite{qi2017pointnet,qi2017pointnet++,dgcnn} adopt hierarchical segmentation heads to progressively upsample the point features from intermediate layers, while our head contains no hierarchical structure and only relies on the pre-trained encoder to capture the multi-scale information of point clouds. The results fully demonstrate the significance of Point-M2AE's multi-scale pre-training to segmentation tasks.

\begin{table}[t!]
\small
\label{fewshot}
\tabcaption{\textbf{Few-shot classification on ModelNet40~\cite{modelnet40}}. We report the average accuracy ($\%$) and standard deviation ($\%$) of 10 independent experiments. 
}
\centering
\begin{adjustbox}{width=0.8\linewidth}
	\begin{tabular}{lc c c c c}
	\toprule
		\makecell*[c]{\multirow{2}*{Method}} &\multicolumn{2}{c}{5-way} &\multicolumn{2}{c}{10-way}\\
		 \cmidrule(lr){2-3} \cmidrule(lr){4-5}
		 &10-shot &20-shot &10-shot &20-shot\\
		 \cmidrule(lr){1-1} \cmidrule(lr){2-5}
		DGCNN~\cite{dgcnn} &91.8\ $\pm$\ 3.7 &93.4\ $\pm$\ 3.2 &86.3\ $\pm$\ 6.2 &90.9\ $\pm$\ 5.1\\
		{[S]} DGCNN + OcCo~\cite{occo} &91.9\ $\pm$\ 3.3 &93.9\ $\pm$\ 3.1 &86.4\ $\pm$\ 5.4 &91.3\ $\pm$\ 4.6\\
		\cmidrule(lr){1-5}
	    Transformer~\cite{pointbert} &87.8\ $\pm$\ 5.2 &93.3\ $\pm$\ 4.3 &84.6\ $\pm$\ 5.5 &89.4\ $\pm$\ 6.3\\
		{[S]} Transformer + OcCo~\cite{pointbert} &94.0\ $\pm$\ 3.6 &95.9\ $\pm$\ 2.3 &89.4\ $\pm$\ 5.1 &92.4\ $\pm$\ 4.6\\
		{[S]} Point-BERT~\cite{pointbert} &94.6\ $\pm$\ 3.1 &96.3\ $\pm$\ 2.7 &91.0\ $\pm$\ 5.4 &92.7\ $\pm$\ 5.1\vspace{0.05cm}\\
	    \rowcolor{gray!8} \textbf{[S] Point-M2AE} &\textbf{96.8\ $\pm$\ 1.8}& \textbf{98.3\ $\pm$\ 1.4}&\textbf{92.3\ $\pm$\ 4.5} &\textbf{95.0\ $\pm$\ 3.0}\vspace{0.1cm}\\
	    \textit{Improvement} &\textcolor{blue}{+2.2} &\textcolor{blue}{+2.0} &\textcolor{blue}{+1.3} &\textcolor{blue}{+2.3}\\
	\bottomrule
	\end{tabular}
\end{adjustbox}
\vspace*{-0.5cm}
\end{table}

\begin{figure*}[t!]
\vspace{0.2cm}
\begin{minipage}[h!]{0.53\linewidth}
\vspace{0.3cm}
\centering
\small
\label{seg}
\tabcaption{\textbf{Part segmentation on ShapeNetPart~\cite{shapenetpart}}. `mIoU$_C$' (\%) and `mIoU$_I$' (\%) denote the mean IoU across all part categories and all instances in the dataset, respectively.}
\vspace{0.2cm}
\begin{adjustbox}{width=0.95\linewidth}
	\begin{tabular}{lccc}
	\toprule
		Method &mIoU$_C$ &mIoU$_I$\\
		\cmidrule(lr){1-1} \cmidrule(lr){2-2} \cmidrule(lr){3-3}  
	    PointNet~\cite{qi2017pointnet}  &80.39 &83.70 \\
	    PointNet++~\cite{qi2017pointnet++} &81.85 &85.10 \\
	    DGCNN~\cite{dgcnn} &82.33 &85.20 \\
	    \cmidrule(lr){1-3}
	    Transformer~\cite{pointbert} &83.42 &85.10 \\
	    {[S]} Transformer + OcCo~\cite{pointbert} &83.42 &85.10 \\
	    {[S]} Point-BERT~\cite{pointbert} &84.11 &85.60 \\
	    \rowcolor{gray!8} \textbf{[S] Point-M2AE} &\textbf{84.86} &\textbf{86.51}\vspace{0.1cm}\\
	    \textit{Improvement} &\textcolor{blue}{+0.75} &\textcolor{blue}{+0.91} \\
	  \bottomrule
	\end{tabular}
\end{adjustbox}
\end{minipage}\quad
\begin{minipage}[h!]{0.43\linewidth}
\centering
\small
\vspace{0.3cm}
\label{detect}
\tabcaption{\textbf{3D object detection on ScanNetV2~\cite{ScanNetV2}.} We report the performance (\%) of self-supervised learning methods based on VoteNet~\cite{votenet} and 3DETR-m~\cite{3detr}.}
\vspace{0.2cm}
\begin{adjustbox}{width=0.95\linewidth}
		\begin{tabular}{lcc}
	\toprule
		Method &AP$_{25}$ &AP$_{50}$\\
		\cmidrule(lr){1-1} \cmidrule(lr){2-2}\cmidrule(lr){3-3}
		VoteNet~\cite{votenet} &58.6 &33.5 \\
	    {[S]} STRL~\cite{strl} &59.5 &38.4 \\
	    {[S]} PointContrast~\cite{pointcontrast}  &59.2 &38.0 \\
	    {[S]} DepthContrast~\cite{depthcontrast}  &61.3 &– \\
	    \cmidrule(lr){1-3}
	   3DETR~\cite{3detr} &62.1 &37.9 \\
	   3DETR-m~\cite{3detr} &65.0 &47.0 \\
	    \rowcolor{gray!8} \textbf{[S] Point-M2AE} &\textbf{66.3}  &\textbf{48.3}\vspace{0.1cm} \\
	    \textit{Improvement} &\textcolor{blue}{+1.3} &\textcolor{blue}{+1.3} \\
	  \bottomrule
	\end{tabular}
\end{adjustbox}
\end{minipage}
\vspace*{-0.1cm}
\end{figure*}

\paragraph{Few-shot Classification.}
We conduct experiments for few-shot classification on ModelNet40~\cite{modelnet40} to evaluate the performance of Point-M2AE with limited fine-tuning data. 
As reported in Table~\ref{fewshot}, Point-M2AE achieves the best performance for all four settings, and surpasses Point-BERT by +2.2\%, +2.0\%, +1.3\%, and +2.7\%, respectively. Our approach also shows smaller deviations than other transformer-based methods, which indicates Point-M2AE has learned to produce more universal 3D representations for well adapting to downstream tasks under low-data regimes.

\paragraph{3D Object Detection}
To further evaluate our hierarchical pre-training on 3D object detection, we apply Point-M2AE to serving as the feature backbone on the indoor ScanNetV2~\cite{ScanNetV2} dataset. We select 3DETR-m~\cite{3detr} as our baseline, which consists of a 3-block encoder and a transformer decoder. Considering the quite different dataset statistics, e.g., 2k input points for ShapeNet~\cite{chang2015shapenet} and 50k input points for ScanNetV2, we adopt the same encoder architecture with that of 3DETR-m, and keep our hierarchical decoder with skip connections unchanged for self-supervised pre-training on ScanNetV2.
More details of models and training are in Appendix. As reported in Table~\ref{detect}, compared to training from scratch, our hierarchical pre-training boosts the performance of 3DETR-m by +1.34\% AP$_{25}$ and +1.29\% AP$_{50}$. The experiments demonstrate the effectiveness of Point-M2AE to learn multi-scale point cloud encoding for object detection and its potential to benefit a wider range of 3D applications.

\begin{figure*}[t!]
  \centering
    \includegraphics[width=\textwidth]{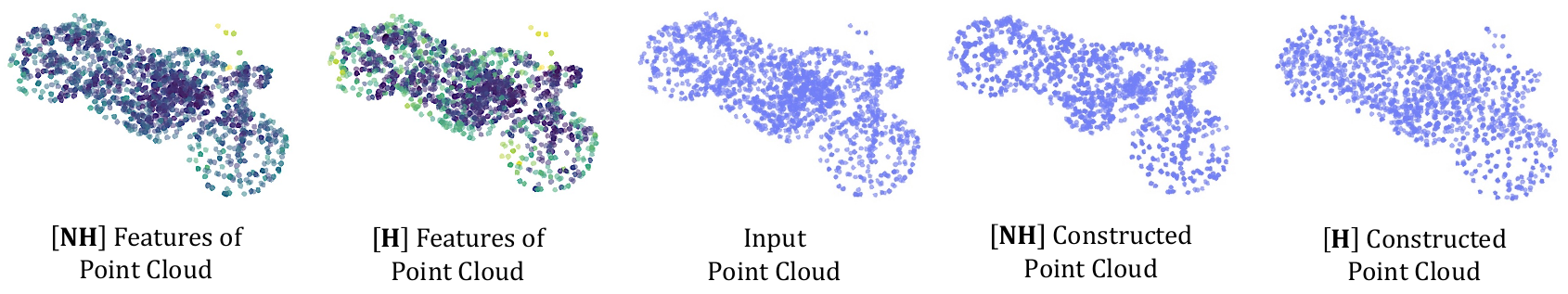}
    \vspace{-0.2cm}
   \figcaption{\textbf{Visualization of fine-grained information.} We denote the outputs from hierarchical and non-hierarchical architectures as \textbf{[NH]} and \textbf{[H]}, respectively.
   For an input point cloud (Middle), we visualize its extracted features (Left) and reconstruction results (Right).}
    \label{f6}
\vspace{-0.2cm}
\end{figure*}

\subsection{Ablation Study}
\label{ablation}
 We conduct ablation study by modifying one of the components at a time during pre-training and explore the best masking strategy. We report the classification accuracy on ModelNet40~\cite{modelnet40} by linear SVM to evaluate the pre-trained representations. For downstream tasks, we train the network from scratch to validate the significance of our hierarchical pre-training.

\begin{figure*}[t!]
\begin{minipage}[t!]{0.5\linewidth}
\centering
\small
\tabcaption{\textbf{Hierarchical Modules.}
`H' represents the encoder and decoder with multi-stage hierarchies. `Skip C.' denotes the skip connections.}
\vspace{0.3cm}
\label{Hmodules}
\begin{adjustbox}{width=0.95\linewidth}
\begin{tabular}{cccc}
\toprule
Encoder & Decoder & Skip C. &Acc. (\%)\\
\cmidrule(lr){1-3} \cmidrule(lr){4-4}
 \rowcolor{gray!8} H &H &\checkmark  &\textbf{92.9}\vspace{0.05cm}\\
- &- &\checkmark  &90.7\vspace{0.05cm}\\
- &H &\checkmark  &91.5\vspace{0.05cm}\\
H &- &\checkmark  &92.2\vspace{0.05cm}\\
H &H &-  &92.1\vspace{0.05cm}\\
\bottomrule
\end{tabular}
\end{adjustbox}
\end{minipage}\quad
\begin{minipage}[t!]{0.47\linewidth}
\centering
\small
\tabcaption{\textbf{Different Masking Strategy.} 
`MS Mask' and `Ratio' denote the multi-scale masking and the mask ratio.}
\vspace{0.3cm}
\label{maskratio}
\begin{adjustbox}{width=0.74\linewidth}
\begin{tabular}{ccc}
\toprule
MS Mask & Ratio &Acc. (\%) \\
\cmidrule(lr){1-2} \cmidrule(lr){3-3}
 \rowcolor{gray!8} \checkmark &0.8 &\textbf{92.9}\vspace{0.05cm}\\
- &0.8 &88.4\vspace{0.05cm}\\
\checkmark &0.6 &92.3\vspace{0.05cm}\\
\checkmark &0.7 &92.7\vspace{0.05cm}\\
\checkmark &0.9 &92.5\vspace{0.05cm}\\
\bottomrule
\end{tabular}
\end{adjustbox}
\end{minipage}
\vspace{-0.2cm}
\end{figure*}

\paragraph{Hierarchical Modules.}
As reported in Table~\ref{Hmodules}, on top of our final solution, Point-M2AE, in the first row, we respectively experiment with removing the hierarchical encoder, hierarchical decoder, and skip connections from our framework. Specifically, we replace our encoder and decoder with 1-stage plain architectures similar to MAE, which contains 15 and 2 vanilla transformer blocks, respectively. We observe the absence of multi-stage structures either in encoder or decoder hurts the performance, and the hierarchical encoder plays a better role than the decoder. Also, the skip connections well benefits the accuracy by providing complementary information for the decoder.

\paragraph{Masking Strategy.}
In Table~\ref{maskratio}, we report Point-M2AE with different mask settings. Without the multi-scale masking, we randomly generate masks at each scale, which leads to fragmented visible regions for all scales. With this strategy, the network would `peek' different parts of the point cloud at different stages, which disturbs the representation learning and harms the performance by -4.5$\%$ accuracy. For different mask ratios, we find the 80$\%$ ratio performs the best to build a properly challenging pretext task for self-supervised pre-training.

\paragraph{With and Without Pre-training.}
\begin{wraptable}{r}{7cm}
\centering
\small
\vspace{-0.8cm}
\tabcaption{\textbf{With and without the pre-training.} `ModelNet40-FS' denotes the few-shot classification on 10-way 20-shot ModelNet40~\cite{modelnet40}.}
\vspace{0.2cm}
\begin{adjustbox}{width=0.92\linewidth}
	\begin{tabular}{lcc}
    \toprule
    \ Dataset &w/o (\%) &w (\%)\\
    \cmidrule(lr){1-1} \cmidrule(lr){2-2} \cmidrule(lr){3-3}
     \ ModelNet40~\cite{modelnet40} &92.5 &94.0\vspace{0.05cm} \\
     \ ScanObjectNN~\cite{scanobjectnn} &83.9 &86.4\vspace{0.05cm} \\
     \ ModelNet40-FS~\cite{modelnet40} &91.2 &95.0\vspace{0.05cm} \\
     \ ShapeNetPart~\cite{shapenetpart} &85.4 &86.5\\
    \bottomrule
\end{tabular}
\end{adjustbox}
\label{wwo}
\end{wraptable}
We report the performance of Point-M2AE on downstream tasks with and without the pre-training in Table~\ref{wwo}. For `w/o', we randomly initialize the parameters and train the network from scratch. As shown, the pre-training can largely boost the performance on four datasets respectively by +1.5$\%$, +2.5$\%$, +3.8$\%$, and +1.1$\%$, which indicates the superiority and significance of our hierarchical pre-training. 

\begin{figure*}[t!]
  \centering
    \includegraphics[width=\textwidth]{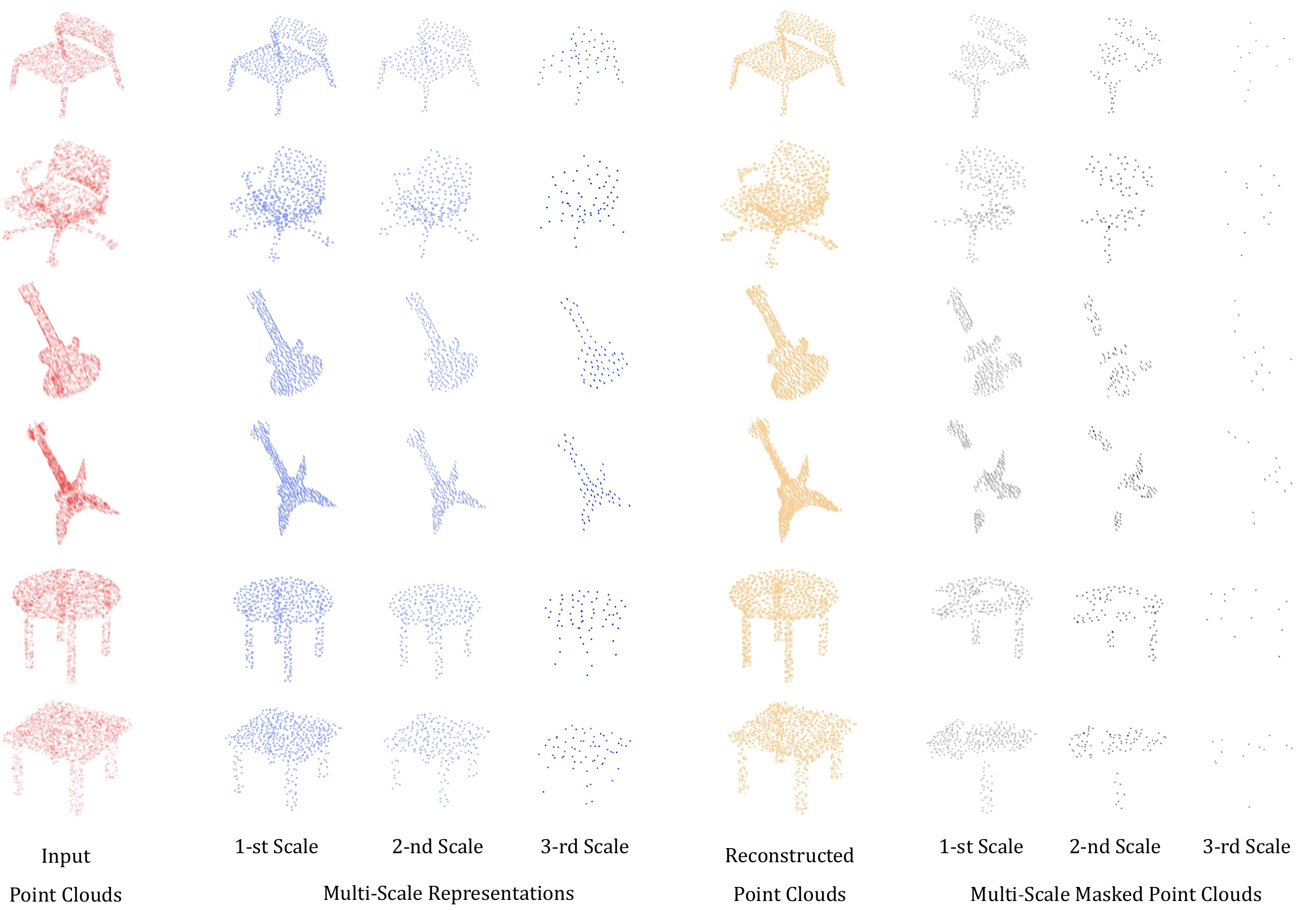}
    \vspace{-0.2cm}
   \figcaption{\textbf{Visualization of multi-scale point clouds.} In each row, we visualize the input point clouds, their multi-scale representations, the reconstructed coordinates, and multi-scale masked point clouds.}
    \label{vis}
\vspace{-0.2cm}
\end{figure*}

\section{Visualization}
\paragraph{Multi-scale Masking.}
To ease the understanding of our multi-scale masking strategy, we visualize the input point cloud, the 3-scale representations, the reconstructed point cloud, and 3-scale masked point clouds, respectively in each row of Figure~\ref{vis}. As shown, different scales can represent different levels of geometric details and semantics for point clouds. By the multi-scale masking strategy, we observe the visible positions of masked point clouds are block-wise within one scale and consistent across scales, which is significant for our hierarchical pre-training.
\vspace{-0.1cm}
\paragraph{Fine-grained Information.}
The fine-grained 3D structures, e.g., thin branches of a plant, fingers of a human, engines of a plane, are significant to distinguish similar shapes and can be well encoded by our hierarchical representations. In Figure~\ref{f6}, we compare our Point-M2AE with multi-stage, [H], and single-scale, [NH], architectures by visualizing their extracted point features and reconstructed point clouds during pre-training. In contarst to the single-scale network, the multi-scale one indicates higher feature responses in the fine-grained structures and reconstructs more accurate spatial details.

\vspace{-0.1cm}
\section{Conclusion}
We propose Point-M2AE, a multi-scale masked autoencoder for self-supervised pre-training on 3D point clouds. With a hierarchical architecture, Point-M2AE learns to produce powerful 3D representations by encoding multi-scale point clouds and reconstructing the masked coordinates from a global-to-local upsampling scheme. Extensive experiments have demonstrated the superiority of Point-M2AE to be a strong 3D representation learner.
For limitations and future work, we will focus on applying Point-M2AE for wider 3D applications, e.g., outdoor and open-world scene understanding. We do not foresee negative social impact from the proposed work.

\textbf{Acknowledgement.} This work is supported by the National Natural Science Foundation of China (Grant No. 62206272), Shanghai Committee of Science and Technology (Grant No. 21DZ1100100), Centre for Perceptual and Interactive Intelligence Limited, and the General Research Fund through the Research Grants Council of Hong Kong (Grant No. 14204021, 14207319).

{
\bibliographystyle{bib/ieee_fullname}
\bibliography{bib/egbib}

\begin{thebibliography}{10}\itemsep=-1pt

\bibitem{achituve2021self}
Idan Achituve, Haggai Maron, and Gal Chechik.
\newblock Self-supervised learning for domain adaptation on point clouds.
\newblock In {\em Proceedings of the IEEE/CVF Winter Conference on Applications
  of Computer Vision}, pages 123--133, 2021.

\bibitem{afham2022crosspoint}
Mohamed Afham, Isuru Dissanayake, Dinithi Dissanayake, Amaya Dharmasiri,
  Kanchana Thilakarathna, and Ranga Rodrigo.
\newblock Crosspoint: Self-supervised cross-modal contrastive learning for 3d
  point cloud understanding.
\newblock {\em arXiv preprint arXiv:2203.00680}, 2022.

\bibitem{baevski2022data2vec}
Alexei Baevski, Wei-Ning Hsu, Qiantong Xu, Arun Babu, Jiatao Gu, and Michael
  Auli.
\newblock Data2vec: A general framework for self-supervised learning in speech,
  vision and language.
\newblock {\em arXiv preprint arXiv:2202.03555}, 2022.

\bibitem{bao2021beit}
Hangbo Bao, Li Dong, and Furu Wei.
\newblock Beit: Bert pre-training of image transformers.
\newblock {\em arXiv preprint arXiv:2106.08254}, 2021.

\bibitem{gpt3}
Tom Brown, Benjamin Mann, Nick Ryder, Melanie Subbiah, Jared~D Kaplan, Prafulla
  Dhariwal, Arvind Neelakantan, Pranav Shyam, Girish Sastry, Amanda Askell,
  et~al.
\newblock Language models are few-shot learners.
\newblock {\em Advances in neural information processing systems},
  33:1877--1901, 2020.

\bibitem{carion2020end}
Nicolas Carion, Francisco Massa, Gabriel Synnaeve, Nicolas Usunier, Alexander
  Kirillov, and Sergey Zagoruyko.
\newblock End-to-end object detection with transformers.
\newblock In {\em European conference on computer vision}, pages 213--229.
  Springer, 2020.

\bibitem{chang2015shapenet}
Angel~X Chang, Thomas Funkhouser, Leonidas Guibas, Pat Hanrahan, Qixing Huang,
  Zimo Li, Silvio Savarese, Manolis Savva, Shuran Song, Hao Su, et~al.
\newblock Shapenet: An information-rich 3d model repository.
\newblock {\em arXiv preprint arXiv:1512.03012}, 2015.

\bibitem{chen2020simple}
Ting Chen, Simon Kornblith, Mohammad Norouzi, and Geoffrey Hinton.
\newblock A simple framework for contrastive learning of visual
  representations.
\newblock In {\em International conference on machine learning}, pages
  1597--1607. PMLR, 2020.

\bibitem{chen2021exploring}
Xinlei Chen and Kaiming He.
\newblock Exploring simple siamese representation learning.
\newblock In {\em Proceedings of the IEEE/CVF Conference on Computer Vision and
  Pattern Recognition}, pages 15750--15758, 2021.

\bibitem{ScanNetV2}
Angela Dai, Angel~X Chang, Manolis Savva, Maciej Halber, Thomas Funkhouser, and
  Matthias Nie{\ss}ner.
\newblock Scannet: Richly-annotated 3d reconstructions of indoor scenes.
\newblock In {\em Proceedings of the IEEE conference on computer vision and
  pattern recognition}, pages 5828--5839, 2017.

\bibitem{bert}
Jacob Devlin, Ming-Wei Chang, Kenton Lee, and Kristina Toutanova.
\newblock Bert: Pre-training of deep bidirectional transformers for language
  understanding.
\newblock {\em arXiv preprint arXiv:1810.04805}, 2018.

\bibitem{devlin2018bert}
Jacob Devlin, Ming-Wei Chang, Kenton Lee, and Kristina Toutanova.
\newblock Bert: Pre-training of deep bidirectional transformers for language
  understanding.
\newblock {\em arXiv preprint arXiv:1810.04805}, 2018.

\bibitem{votenet}
Zhipeng Ding, Xu Han, and Marc Niethammer.
\newblock Votenet: A deep learning label fusion method for multi-atlas
  segmentation.
\newblock In {\em International Conference on Medical Image Computing and
  Computer-Assisted Intervention}, pages 202--210. Springer, 2019.

\bibitem{vit}
Alexey Dosovitskiy, Lucas Beyer, Alexander Kolesnikov, Dirk Weissenborn,
  Xiaohua Zhai, Thomas Unterthiner, Mostafa Dehghani, Matthias Minderer, Georg
  Heigold, Sylvain Gelly, et~al.
\newblock An image is worth 16x16 words: Transformers for image recognition at
  scale.
\newblock {\em arXiv preprint arXiv:2010.11929}, 2020.

\bibitem{chamfer}
Haoqiang Fan, Hao Su, and Leonidas~J Guibas.
\newblock A point set generation network for 3d object reconstruction from a
  single image.
\newblock In {\em Proceedings of the IEEE conference on computer vision and
  pattern recognition}, pages 605--613, 2017.

\bibitem{gao2021fast}
Peng Gao, Minghang Zheng, Xiaogang Wang, Jifeng Dai, and Hongsheng Li.
\newblock Fast convergence of detr with spatially modulated co-attention.
\newblock In {\em Proceedings of the IEEE/CVF International Conference on
  Computer Vision}, pages 3621--3630, 2021.

\bibitem{guo2021pct}
Meng-Hao Guo, Jun-Xiong Cai, Zheng-Ning Liu, Tai-Jiang Mu, Ralph~R Martin, and
  Shi-Min Hu.
\newblock Pct: Point cloud transformer.
\newblock {\em Computational Visual Media}, 7(2):187--199, 2021.

\bibitem{vipgan}
Zhizhong Han, Mingyang Shang, Yu-Shen Liu, and Matthias Zwicker.
\newblock View inter-prediction gan: Unsupervised representation learning for
  3d shapes by learning global shape memories to support local view
  predictions.
\newblock In {\em Proceedings of the AAAI Conference on Artificial
  Intelligence}, volume~33, pages 8376--8384, 2019.

\bibitem{mapvae}
Zhizhong Han, Xiyang Wang, Yu-Shen Liu, and Matthias Zwicker.
\newblock Multi-angle point cloud-vae: Unsupervised feature learning for 3d
  point clouds from multiple angles by joint self-reconstruction and
  half-to-half prediction.
\newblock In {\em 2019 IEEE/CVF International Conference on Computer Vision
  (ICCV)}, pages 10441--10450. IEEE, 2019.

\bibitem{mae}
Kaiming He, Xinlei Chen, Saining Xie, Yanghao Li, Piotr Doll{\'a}r, and Ross
  Girshick.
\newblock Masked autoencoders are scalable vision learners.
\newblock {\em arXiv preprint arXiv:2111.06377}, 2021.

\bibitem{he2020momentum}
Kaiming He, Haoqi Fan, Yuxin Wu, Saining Xie, and Ross Girshick.
\newblock Momentum contrast for unsupervised visual representation learning.
\newblock In {\em Proceedings of the IEEE/CVF conference on computer vision and
  pattern recognition}, pages 9729--9738, 2020.

\bibitem{strl}
Siyuan Huang, Yichen Xie, Song-Chun Zhu, and Yixin Zhu.
\newblock Spatio-temporal self-supervised representation learning for 3d point
  clouds.
\newblock In {\em Proceedings of the IEEE/CVF International Conference on
  Computer Vision}, pages 6535--6545, 2021.

\bibitem{align}
Chao Jia, Yinfei Yang, Ye Xia, Yi-Ting Chen, Zarana Parekh, Hieu Pham, Quoc Le,
  Yun-Hsuan Sung, Zhen Li, and Tom Duerig.
\newblock Scaling up visual and vision-language representation learning with
  noisy text supervision.
\newblock In {\em International Conference on Machine Learning}, pages
  4904--4916. PMLR, 2021.

\bibitem{kingma2014adam}
Diederik~P Kingma and Jimmy Ba.
\newblock Adam: A method for stochastic optimization.
\newblock {\em arXiv preprint arXiv:1412.6980}, 2014.

\bibitem{sonet}
Jiaxin Li, Ben~M Chen, and Gim~Hee Lee.
\newblock So-net: Self-organizing network for point cloud analysis.
\newblock In {\em Proceedings of the IEEE conference on computer vision and
  pattern recognition}, pages 9397--9406, 2018.

\bibitem{li2022uniformer}
Kunchang Li, Yali Wang, Junhao Zhang, Peng Gao, Guanglu Song, Yu Liu, Hongsheng
  Li, and Yu Qiao.
\newblock Uniformer: Unifying convolution and self-attention for visual
  recognition.
\newblock {\em arXiv preprint arXiv:2201.09450}, 2022.

\bibitem{li2018pointcnn}
Yangyan Li, Rui Bu, Mingchao Sun, Wei Wu, Xinhan Di, and Baoquan Chen.
\newblock Pointcnn: Convolution on x-transformed points.
\newblock {\em Advances in neural information processing systems}, 31:820--830,
  2018.

\bibitem{densepoint}
Yongcheng Liu, Bin Fan, Gaofeng Meng, Jiwen Lu, Shiming Xiang, and Chunhong
  Pan.
\newblock Densepoint: Learning densely contextual representation for efficient
  point cloud processing.
\newblock In {\em Proceedings of the IEEE/CVF International Conference on
  Computer Vision}, pages 5239--5248, 2019.

\bibitem{rscnn}
Yongcheng Liu, Bin Fan, Shiming Xiang, and Chunhong Pan.
\newblock Relation-shape convolutional neural network for point cloud analysis.
\newblock In {\em Proceedings of the IEEE/CVF Conference on Computer Vision and
  Pattern Recognition}, pages 8895--8904, 2019.

\bibitem{liu2021swin}
Ze Liu, Yutong Lin, Yue Cao, Han Hu, Yixuan Wei, Zheng Zhang, Stephen Lin, and
  Baining Guo.
\newblock Swin transformer: Hierarchical vision transformer using shifted
  windows.
\newblock In {\em Proceedings of the IEEE/CVF International Conference on
  Computer Vision}, pages 10012--10022, 2021.

\bibitem{mao2021dual}
Mingyuan Mao, Renrui Zhang, Honghui Zheng, Teli Ma, Yan Peng, Errui Ding,
  Baochang Zhang, Shumin Han, et~al.
\newblock Dual-stream network for visual recognition.
\newblock {\em Advances in Neural Information Processing Systems}, 34, 2021.

\bibitem{3detr}
Ishan Misra, Rohit Girdhar, and Armand Joulin.
\newblock An end-to-end transformer model for 3d object detection.
\newblock In {\em Proceedings of the IEEE/CVF International Conference on
  Computer Vision (ICCV)}, pages 2906--2917, October 2021.

\bibitem{pang2022masked}
Yatian Pang, Wenxiao Wang, Francis~EH Tay, Wei Liu, Yonghong Tian, and Li Yuan.
\newblock Masked autoencoders for point cloud self-supervised learning.
\newblock {\em arXiv preprint arXiv:2203.06604}, 2022.

\bibitem{poursaeed2020self}
Omid Poursaeed, Tianxing Jiang, Han Qiao, Nayun Xu, and Vladimir~G Kim.
\newblock Self-supervised learning of point clouds via orientation estimation.
\newblock In {\em 2020 International Conference on 3D Vision (3DV)}, pages
  1018--1028. IEEE, 2020.

\bibitem{qi2017pointnet}
Charles~R Qi, Hao Su, Kaichun Mo, and Leonidas~J Guibas.
\newblock Pointnet: Deep learning on point sets for 3d classification and
  segmentation.
\newblock In {\em Proceedings of the IEEE conference on computer vision and
  pattern recognition}, pages 652--660, 2017.

\bibitem{qi2017pointnet++}
Charles~R Qi, Li Yi, Hao Su, and Leonidas~J Guibas.
\newblock Pointnet++: Deep hierarchical feature learning on point sets in a
  metric space.
\newblock {\em arXiv preprint arXiv:1706.02413}, 2017.

\bibitem{clip}
Alec Radford, Jong~Wook Kim, Chris Hallacy, Aditya Ramesh, Gabriel Goh,
  Sandhini Agarwal, Girish Sastry, Amanda Askell, Pamela Mishkin, Jack Clark,
  et~al.
\newblock Learning transferable visual models from natural language
  supervision.
\newblock In {\em International Conference on Machine Learning}, pages
  8748--8763. PMLR, 2021.

\bibitem{gpt1}
Alec Radford and Karthik Narasimhan.
\newblock Improving language understanding by generative pre-training.
\newblock 2018.

\bibitem{gpt2}
Alec Radford, Jeffrey Wu, Rewon Child, David Luan, Dario Amodei, Ilya
  Sutskever, et~al.
\newblock Language models are unsupervised multitask learners.
\newblock {\em OpenAI blog}, 1(8):9, 2019.

\bibitem{dvae}
Jason~Tyler Rolfe.
\newblock Discrete variational autoencoders.
\newblock {\em arXiv preprint arXiv:1609.02200}, 2016.

\bibitem{unet}
Olaf Ronneberger, Philipp Fischer, and Thomas Brox.
\newblock U-net: Convolutional networks for biomedical image segmentation.
\newblock In {\em International Conference on Medical image computing and
  computer-assisted intervention}, pages 234--241. Springer, 2015.

\bibitem{sauder2019self}
Jonathan Sauder and Bjarne Sievers.
\newblock Self-supervised deep learning on point clouds by reconstructing
  space.
\newblock {\em Advances in Neural Information Processing Systems}, 32, 2019.

\bibitem{jiasaw}
Jonathan Sauder and Bjarne Sievers.
\newblock Self-supervised deep learning on point clouds by reconstructing
  space.
\newblock {\em Advances in Neural Information Processing Systems}, 32, 2019.

\bibitem{touvron2021training}
Hugo Touvron, Matthieu Cord, Matthijs Douze, Francisco Massa, Alexandre
  Sablayrolles, and Herv{\'e} J{\'e}gou.
\newblock Training data-efficient image transformers \& distillation through
  attention.
\newblock In {\em International Conference on Machine Learning}, pages
  10347--10357. PMLR, 2021.

\bibitem{scanobjectnn}
Mikaela~Angelina Uy, Quang-Hieu Pham, Binh-Son Hua, Thanh Nguyen, and Sai-Kit
  Yeung.
\newblock Revisiting point cloud classification: A new benchmark dataset and
  classification model on real-world data.
\newblock In {\em Proceedings of the IEEE/CVF International Conference on
  Computer Vision}, pages 1588--1597, 2019.

\bibitem{latentgan}
Diego Valsesia, Giulia Fracastoro, and Enrico Magli.
\newblock Learning localized representations of point clouds with
  graph-convolutional generative adversarial networks.
\newblock {\em IEEE Transactions on Multimedia}, 23:402--414, 2020.

\bibitem{van2008visualizing}
Laurens Van~der Maaten and Geoffrey Hinton.
\newblock Visualizing data using t-sne.
\newblock {\em Journal of machine learning research}, 9(11), 2008.

\bibitem{vaswani2017attention}
Ashish Vaswani, Noam Shazeer, Niki Parmar, Jakob Uszkoreit, Llion Jones,
  Aidan~N Gomez, {\L}ukasz Kaiser, and Illia Polosukhin.
\newblock Attention is all you need.
\newblock {\em Advances in neural information processing systems}, 30, 2017.

\bibitem{occo}
Hanchen Wang, Qi Liu, Xiangyu Yue, Joan Lasenby, and Matt~J Kusner.
\newblock Unsupervised point cloud pre-training via occlusion completion.
\newblock In {\em Proceedings of the IEEE/CVF International Conference on
  Computer Vision}, pages 9782--9792, 2021.

\bibitem{dgcnn}
Yue Wang, Yongbin Sun, Ziwei Liu, Sanjay~E Sarma, Michael~M Bronstein, and
  Justin~M Solomon.
\newblock Dynamic graph cnn for learning on point clouds.
\newblock {\em Acm Transactions On Graphics (tog)}, 38(5):1--12, 2019.

\bibitem{wei2021masked}
Chen Wei, Haoqi Fan, Saining Xie, Chao-Yuan Wu, Alan Yuille, and Christoph
  Feichtenhofer.
\newblock Masked feature prediction for self-supervised visual pre-training.
\newblock {\em arXiv preprint arXiv:2112.09133}, 2021.

\bibitem{3dgan}
Jiajun Wu, Chengkai Zhang, Tianfan Xue, Bill Freeman, and Josh Tenenbaum.
\newblock Learning a probabilistic latent space of object shapes via 3d
  generative-adversarial modeling.
\newblock {\em Advances in neural information processing systems}, 29, 2016.

\bibitem{modelnet40}
Zhirong Wu, Shuran Song, Aditya Khosla, Fisher Yu, Linguang Zhang, Xiaoou Tang,
  and Jianxiong Xiao.
\newblock 3d shapenets: A deep representation for volumetric shapes.
\newblock In {\em Proceedings of the IEEE conference on computer vision and
  pattern recognition}, pages 1912--1920, 2015.

\bibitem{xie2021segformer}
Enze Xie, Wenhai Wang, Zhiding Yu, Anima Anandkumar, Jose~M Alvarez, and Ping
  Luo.
\newblock Segformer: Simple and efficient design for semantic segmentation with
  transformers.
\newblock {\em Advances in Neural Information Processing Systems}, 34, 2021.

\bibitem{pointcontrast}
Saining Xie, Jiatao Gu, Demi Guo, Charles~R Qi, Leonidas Guibas, and Or Litany.
\newblock Pointcontrast: Unsupervised pre-training for 3d point cloud
  understanding.
\newblock In {\em European conference on computer vision}, pages 574--591.
  Springer, 2020.

\bibitem{xie2021simmim}
Zhenda Xie, Zheng Zhang, Yue Cao, Yutong Lin, Jianmin Bao, Zhuliang Yao, Qi
  Dai, and Han Hu.
\newblock Simmim: A simple framework for masked image modeling.
\newblock {\em arXiv preprint arXiv:2111.09886}, 2021.

\bibitem{foldingnet}
Yaoqing Yang, Chen Feng, Yiru Shen, and Dong Tian.
\newblock Foldingnet: Point cloud auto-encoder via deep grid deformation.
\newblock In {\em Proceedings of the IEEE conference on computer vision and
  pattern recognition}, pages 206--215, 2018.

\bibitem{shapenetpart}
Li Yi, Vladimir~G Kim, Duygu Ceylan, I-Chao Shen, Mengyan Yan, Hao Su, Cewu Lu,
  Qixing Huang, Alla Sheffer, and Leonidas Guibas.
\newblock A scalable active framework for region annotation in 3d shape
  collections.
\newblock {\em ACM Transactions on Graphics (ToG)}, 35(6):1--12, 2016.

\bibitem{yu2021pointr}
Xumin Yu, Yongming Rao, Ziyi Wang, Zuyan Liu, Jiwen Lu, and Jie Zhou.
\newblock Pointr: Diverse point cloud completion with geometry-aware
  transformers.
\newblock In {\em Proceedings of the IEEE/CVF International Conference on
  Computer Vision}, pages 12498--12507, 2021.

\bibitem{pointbert}
Xumin Yu, Lulu Tang, Yongming Rao, Tiejun Huang, Jie Zhou, and Jiwen Lu.
\newblock Point-bert: Pre-training 3d point cloud transformers with masked
  point modeling.
\newblock {\em arXiv preprint arXiv:2111.14819}, 2021.

\bibitem{zhang2021pointclip}
Renrui Zhang, Ziyu Guo, Wei Zhang, Kunchang Li, Xupeng Miao, Bin Cui, Yu Qiao,
  Peng Gao, and Hongsheng Li.
\newblock Pointclip: Point cloud understanding by clip.
\newblock {\em arXiv preprint arXiv:2112.02413}, 2021.

\bibitem{zhang2022monodetr}
Renrui Zhang, Han Qiu, Tai Wang, Xuanzhuo Xu, Ziyu Guo, Yu Qiao, Peng Gao, and
  Hongsheng Li.
\newblock Monodetr: Depth-aware transformer for monocular 3d object detection.
\newblock {\em arXiv preprint arXiv:2203.13310}, 2022.

\bibitem{depthcontrast}
Zaiwei Zhang, Rohit Girdhar, Armand Joulin, and Ishan Misra.
\newblock Self-supervised pretraining of 3d features on any point-cloud.
\newblock In {\em Proceedings of the IEEE/CVF International Conference on
  Computer Vision}, pages 10252--10263, 2021.

\bibitem{pointtransformer}
Hengshuang Zhao, Li Jiang, Jiaya Jia, Philip~HS Torr, and Vladlen Koltun.
\newblock Point transformer.
\newblock In {\em Proceedings of the IEEE/CVF International Conference on
  Computer Vision}, pages 16259--16268, 2021.

\bibitem{zheng2020end}
Minghang Zheng, Peng Gao, Renrui Zhang, Kunchang Li, Xiaogang Wang, Hongsheng
  Li, and Hao Dong.
\newblock End-to-end object detection with adaptive clustering transformer.
\newblock {\em arXiv preprint arXiv:2011.09315}, 2020.

\bibitem{zhou2021ibot}
Jinghao Zhou, Chen Wei, Huiyu Wang, Wei Shen, Cihang Xie, Alan Yuille, and Tao
  Kong.
\newblock ibot: Image bert pre-training with online tokenizer.
\newblock {\em arXiv preprint arXiv:2111.07832}, 2021.

\bibitem{zhu2020deformable}
Xizhou Zhu, Weijie Su, Lewei Lu, Bin Li, Xiaogang Wang, and Jifeng Dai.
\newblock Deformable detr: Deformable transformers for end-to-end object
  detection.
\newblock {\em arXiv preprint arXiv:2010.04159}, 2020.

\end{thebibliography}
}


\section*{Checklist}

\begin{enumerate}

\item For all authors...
\begin{enumerate}
  \item Do the main claims made in the abstract and introduction accurately reflect the paper's contributions and scope?
    \answerYes{}
  \item Did you describe the limitations of your work?
    \answerYes{}
  \item Did you discuss any potential negative societal impacts of your work?
    \answerYes{}
  \item Have you read the ethics review guidelines and ensured that your paper conforms to them?
  \answerYes{}
\end{enumerate}

\item If you are including theoretical results...
\begin{enumerate}
  \item Did you state the full set of assumptions of all theoretical results?
    \answerYes{}
        \item Did you include complete proofs of all theoretical results?
    \answerYes{}
\end{enumerate}

\item If you ran experiments...
\begin{enumerate}
  \item Did you include the code, data, and instructions needed to reproduce the main experimental results (either in the supplemental material or as a URL)?
    \answerYes{}
  \item Did you specify all the training details (e.g., data splits, hyperparameters, how they were chosen)?
    \answerYes{}
        \item Did you report error bars (e.g., with respect to the random seed after running experiments multiple times)?
    \answerYes{}
        \item Did you include the total amount of compute and the type of resources used (e.g., type of GPUs, internal cluster, or cloud provider)?
    \answerYes{}
\end{enumerate}

\item If you are using existing assets (e.g., code, data, models) or curating/releasing new assets...
\begin{enumerate}
  \item If your work uses existing assets, did you cite the creators?
    \answerYes{}
  \item Did you mention the license of the assets?
    \answerNA{}
  \item Did you include any new assets either in the supplemental material or as a URL?
    \answerYes{}
  \item Did you discuss whether and how consent was obtained from people whose data you're using/curating?
    \answerNA{}
  \item Did you discuss whether the data you are using/curating contains personally identifiable information or offensive content?
    \answerNA{}
\end{enumerate}

\item If you used crowdsourcing or conducted research with human subjects...
\begin{enumerate}
  \item Did you include the full text of instructions given to participants and screenshots, if applicable?
    \answerNA{}
  \item Did you describe any potential participant risks, with links to Institutional Review Board (IRB) approvals, if applicable?
    \answerNA{}
  \item Did you include the estimated hourly wage paid to participants and the total amount spent on participant compensation?
    \answerNA{}
\end{enumerate}
\end{enumerate}

\clearpage


\section{Appendix}

\begin{figure*}[h]
  \centering
    \includegraphics[width=\textwidth]{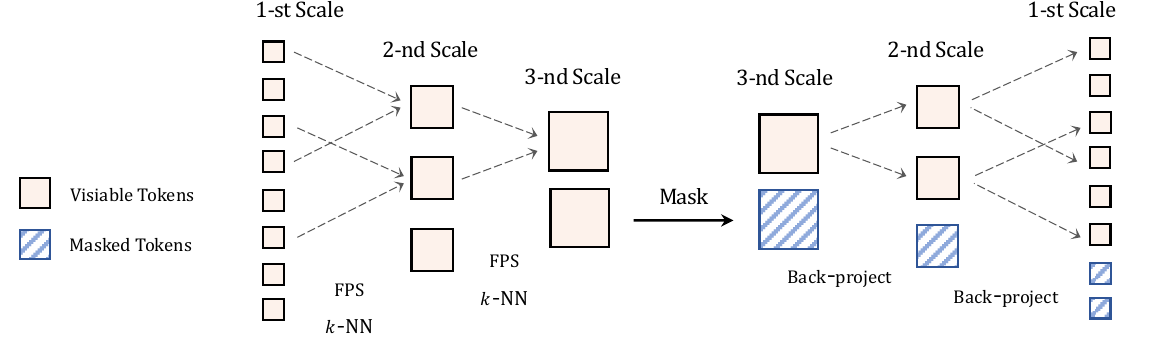}
    \vspace{-0.1cm}
   \figcaption{\textbf{Pipeline of the multi-scale masking.} We first obtain the multi-scale representation of input point clouds by FPS and $k$-NN. Then, we random mask the points at the highest level and back-project the visible positions into precedent scales.}
    \label{pipline}
\end{figure*}

\subsection{Additional Related Work}
\paragraph{Transformers.}
Transformers~\cite{vaswani2017attention} are first proposed in natural language processing to capture the inter-word relations in a long sentence, and have dominated most language tasks~\cite{bert,gpt1,gpt2,gpt3}. Motivated by this, Vision Transformers~\cite{vit} and DETR~\cite{carion2020end} introduce the transformer architecture into computer vision, and stimulate follow-up works to effectively apply transformers to a wide range of vision tasks, such as image classification~\cite{touvron2021training,liu2021swin,mao2021dual}, object detection~\cite{zhu2020deformable,zheng2020end,gao2021fast}, semantic segmentation~\cite{xie2021segformer} and so on~\cite{li2022uniformer}. For 3D understanding, transformer-based networks are also adopted for shape classification, part segmentation~\cite{guo2021pct,pointtransformer}, 3D object detection from point clouds~\cite{3detr} and monocular images~\cite{zhang2022monodetr}. As a pioneer work, PCT~\cite{guo2021pct} utilizes neighbor embedding layers to aggregate local features and processes the downsampled point clouds by transformer blocks. PoinTr~\cite{yu2021pointr} and Point-BERT~\cite{pointbert} divide point clouds into multiple spatial local patches and utilize standard transformers of plain architectures to encode the patches. On top of that, we propose Point-M2AE with a hierarchical encoder-decoder transformer, which is designed for MAE-style self-supervised point cloud pre-training and can well capture the multi-scale features of point clouds.

\subsection{Implementation Details}

\paragraph{Positional Encodings.}
To complement the 3D spatial information, we apply positional encodings to all attention layers in Point-M2AE. For point tokens $T_i^v$ or $\{H_i^m, H_i^v\}$ at stage $i$, we utilize a two-layer MLP to encode its corresponding 3D coordinates $P_i^v$ or $\{P_i^m, P_i^v\}$ into $C_i$-channel vectors, and element-wisely add them with the token features before feeding into the attention layer.

\paragraph{Self-supervised Pre-training.}
Following previous works~\cite{afham2022crosspoint,occo}, we sample 2,048 points from each 3D shape in ShapeNet~\cite{chang2015shapenet} for pre-training Point-M2AE. We pre-train the network for 300 epochs with a batch size 128 and adopt AdamW~\cite{kingma2014adam} as the optimizer. We set the initial learning rate and the weight decay as 10$^{-4}$ and 5$\times$10$^{-2}$, respectively, and adopt the cosine scheduler along with a 10-epoch warm-up. We utilize the common random scaling and random translation for data augmentation during pre-training. For linear SVM on ModelNet40~\cite{modelnet40}, after the hierarchical encoder, we use both max and average pooling to aggregate the features between point tokens, and sum the two pooled features as the encoded global feature of the point cloud.

\paragraph{Shape Classification.}
We fine-tune Point-M2AE on two datasets for shape classification.
The widely adopted ModelNet40~\cite{modelnet40} consists synthetic 3D shapes of 40 categories, in which 9,843 samples are for training, and the other 2,468 are for validation. The challenging ScanObjectNN~\cite{scanobjectnn} contains 11,416 training and 2,882 validation point clouds of 15 categories, which are captured from the noisy real-world scenes and thus have domain gaps with the pre-trained ShapeNet~\cite{chang2015shapenet} dataset. ScanObjectNN is divided into three splits for evaluation, OBJ-BG, OBJ-ONLY and PB-T50-RS, where PB-T50-RS is the most difficult for recognition.
We respectively sample 1,024 and 2,048 points from each 3D shape of ModelNet40 and ScanObjectNN, and utilize only 3-channel coordinates as inputs. The same training settings are adopted for the two datasets. We fine-tune the network for 300 epochs with a batch size 32, and set the learning rate as 5$\times$10$^{-4}$ with a weight decay 5$\times$10$^{-2}$. For other training hyper-parameters, we keep them the same as the pre-training experiment.

\begin{figure*}[t!]
\vspace{-0.1cm}
\centering
\begin{minipage}[h!]{0.48\linewidth}
\centering
\includegraphics[width=0.95\textwidth]{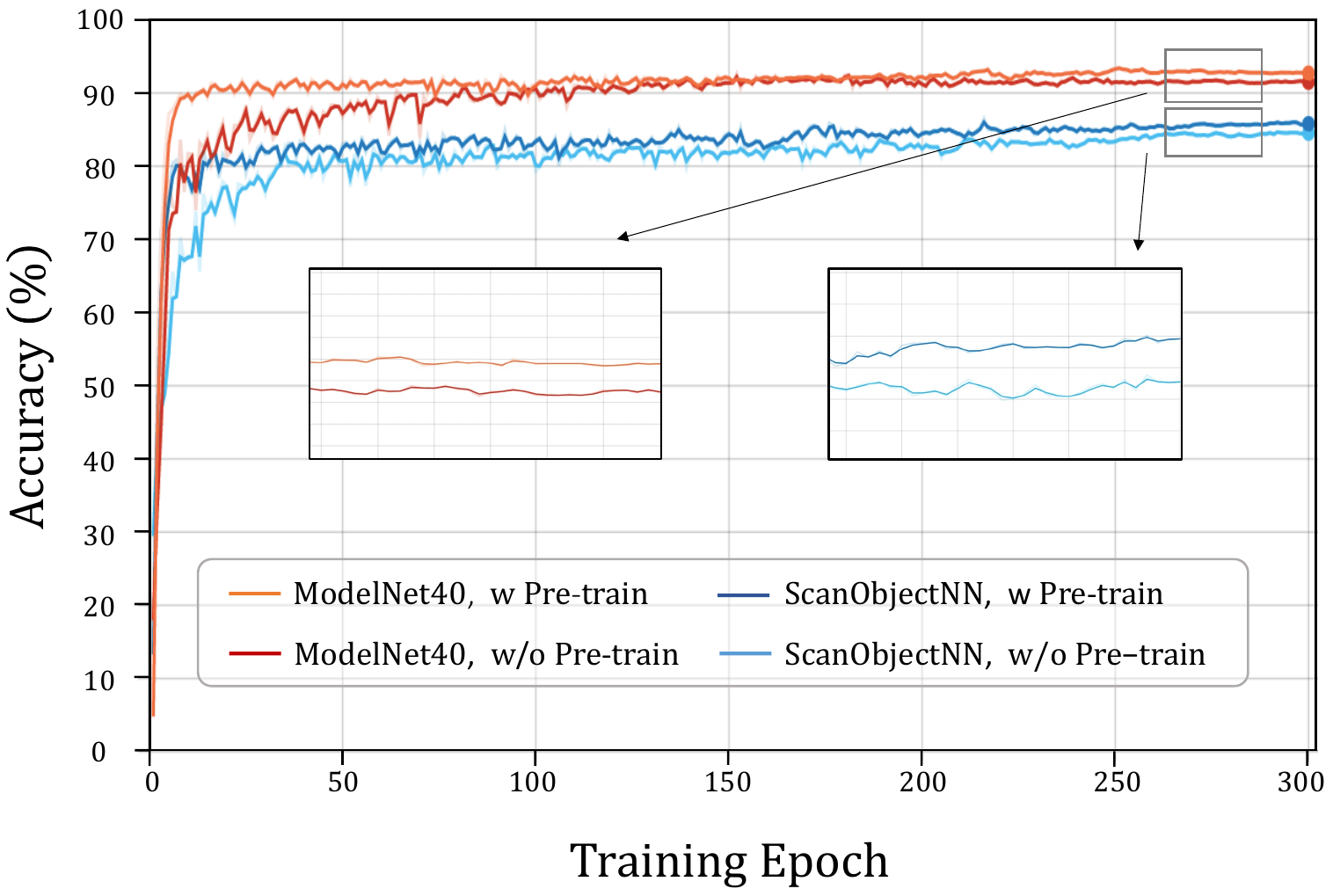}
\end{minipage}\quad
\begin{minipage}[h!]{0.48\linewidth}
\centering
\includegraphics[width=0.95\textwidth]{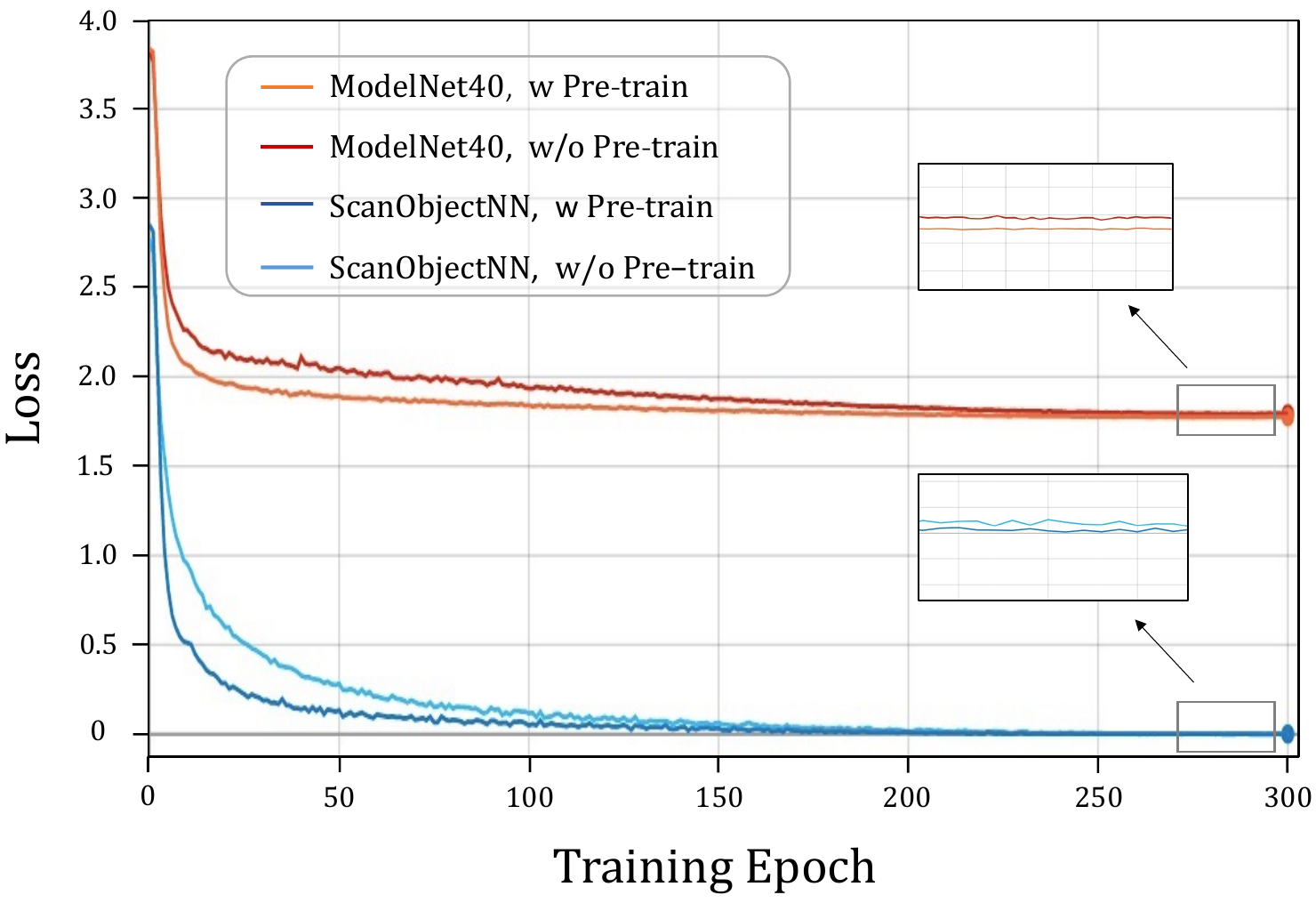}
\end{minipage}
\label{curvef}
\figcaption{\textbf{Learning curves of Point-M2AE with and without pre-training.} We visualize the accuracy (Left) and loss curves (Right) on ModelNet40~\cite{modelnet40} and ScanObjectNN~\cite{scanobjectnn}. We zoom in on the converged accuracy and loss for comparison.}
\end{figure*}

\begin{figure}[t!]
\vspace{0.3cm}
\centering
\includegraphics[width=0.65\textwidth]{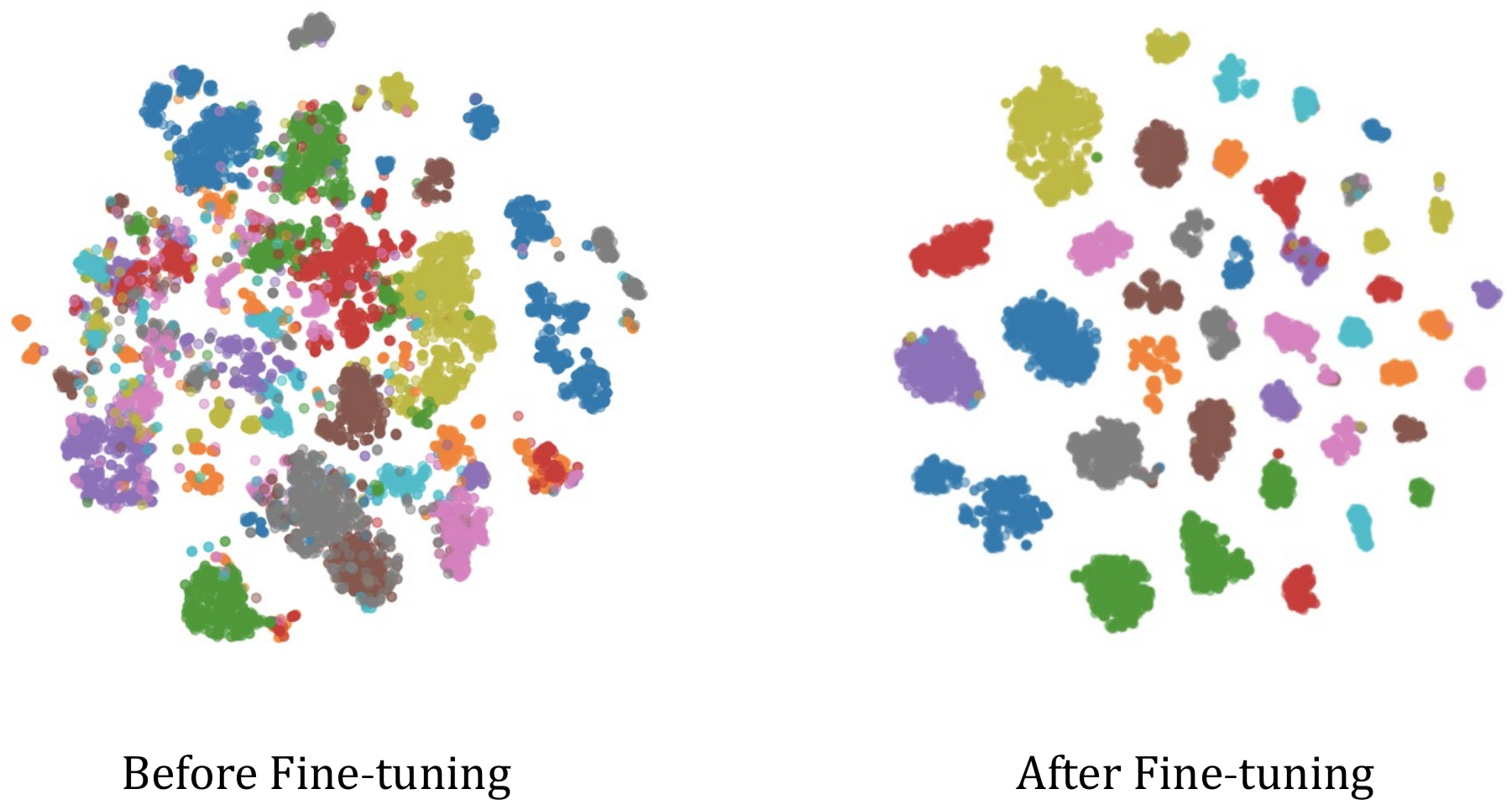}
\figcaption{\textbf{t-SNE~\cite{van2008visualizing} visualization on ModelNet40~\cite{modelnet40}.} We show the features distribution extracted by Point-M2AE before (Left) and after (After) the fine-tuning.}
\vspace{-0.3cm}
\label{tsne}
\end{figure}

\paragraph{Part Segmentation.}
ShapeNetPart~\cite{shapenetpart} contains 16,881 synthetic 3D shapes of 16 object categories and 50 part categories, where 14,007 and 2,874 samples are respectively for training and validation. We sample 2,048 points from each shape as inputs, and predict the part categories for all points. We fine-tune Point-M2AE for 300 epochs with a batch size 16 and set the learning rate as 2$\times$10$^{-4}$ with a weight decay 0.1. Other training settings are the same as the shape classification experiments.

\paragraph{Few-shot Classification.}
We follow previous works~\cite{pointbert,afham2022crosspoint,occo}, to adopt the ``\textit{K}-way \textit{N}-shot'' settings on ModelNet40~\cite{modelnet40} for few-shot classification. We randomly select \textit{K} out of 40 classes and sample $N$+20 3D shapes per class, $N$ for training and 20 for testing. We evaluate Point-M2AE on four few-shot settings: 5-way 10-shot, 5-way 20-shot, 10-way 10-shot, and 10-way 20-shot. To alleviate the variance of random sampling, we conduct 10 independent runs for each few-shot setting and report the average accuracy and standard deviation. We adopt the same training settings as shape classification experiments but only fine-tune Point-M2AE for 150 epochs.

\paragraph{3D Object Detection.}
We pre-train and fine-tune Point-MAE for 3D object detection both on ScanNetV2~\cite{ScanNetV2}. The dataset contains 1,513 scanned indoor scenes with axis-aligned 3D bounding boxes for 18 categories, 1,201 for training and 312 for validation. As we adopt the same encoder architecture in 3DETR-m~\cite{3detr} with 2 stages, we set the stage number of decoder as 1, which accords with the regulation of $S$-stage encoder and $(S-1)$-stage decoder.
We pre-train Point-M2AE for 1,080 epochs with the learning rate 5$\times$10$^{-4}$, and follow other hyper-parameters in the experiment of pre-training on ShapeNet~\cite{chang2015shapenet}. For fine-tuning, we adopt the same settings as training 3DETR-m from scratch in the original paper~\cite{3detr} for fair comparison.


\subsection{Additional Visualization}

\paragraph{Multi-scale Masking Pipeline.}
In Figure~\ref{pipline}, we show the simplified masking pipeline, which clearly illustrates how the mask is generated at the highest scale and back-projects to precedent layers.

\paragraph{Learning Curves.}
To compare the training with and without pre-training, we present their loss and accuracy curves on ModelNet40~\cite{modelnet40} and ScanObjectNN~\cite{scanobjectnn}. As shown in Figure~\ref{curvef}, the curves with pre-training converge faster and achieve higher classification accuracy than the curves without pre-training. This fully demonstrates the effectiveness of Point-M2AE's hierarchical pre-training.

\paragraph{t-SNE~\cite{van2008visualizing}.}
In Figure~\ref{tsne}, we visualize the features distribution extracted by Point-M2AE before and after fine-tuning on ModelNet40~\cite{modelnet40}. As shown, Point-M2AE right after pre-training can already produce discriminative features for different categories without fine-tuning. Then, the fine-tuning further clusters the features of the same category and separates those of different categories.

\begin{figure}[t!]
  \centering
    \includegraphics[width=0.6\textwidth]{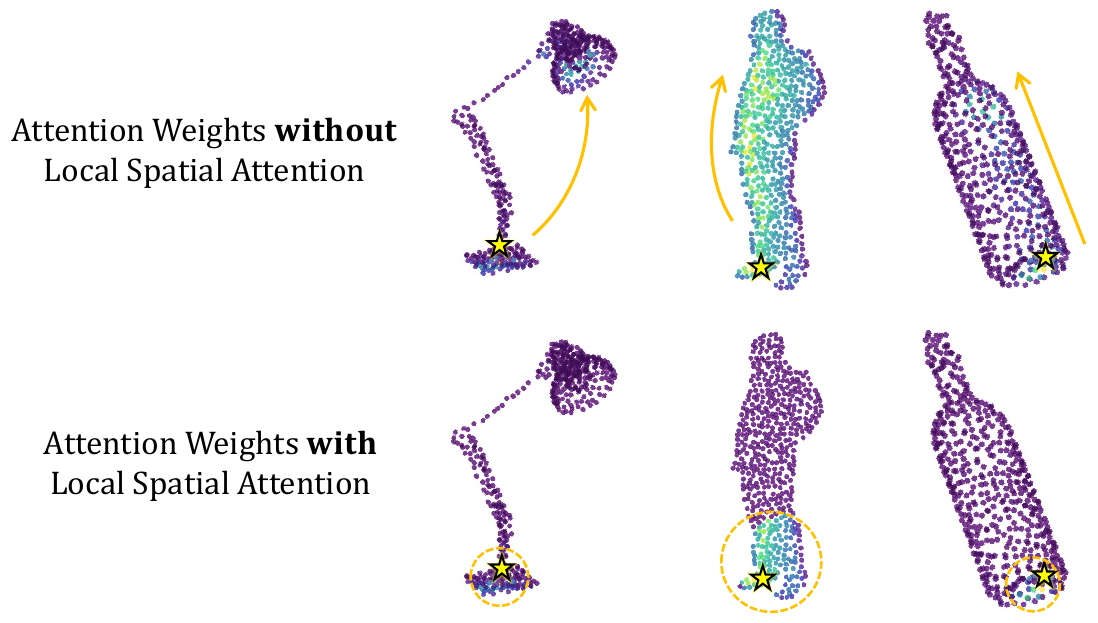}
    \vspace{-0.1cm}
   \figcaption{\textbf{Visualization of local spatial attention.} We visualize the attention weights without (Top) and with (Bottom) local spatial attention. The query points are marked by stars. The attention scopes are marked by arrows and dotted circles in yellow.}
    \label{f5}
\end{figure}

\paragraph{Local Spatial Attention.}
We visualize the attention weights with and without the local attention on ModelNet40~\cite{modelnet40} in Figure~\ref{f5}. As shown, with the local attention, the query point (marked by star) only has large attention values within a local spatial range (marked by yellow dotted circles), other than scattering over the entire 3D shape (marked by yellow arrows). This enables each point to concentrate more on neighboring local features in early stages for capturing detailed structures.

\begin{figure}[t]
  \centering
    \includegraphics[width=\textwidth]{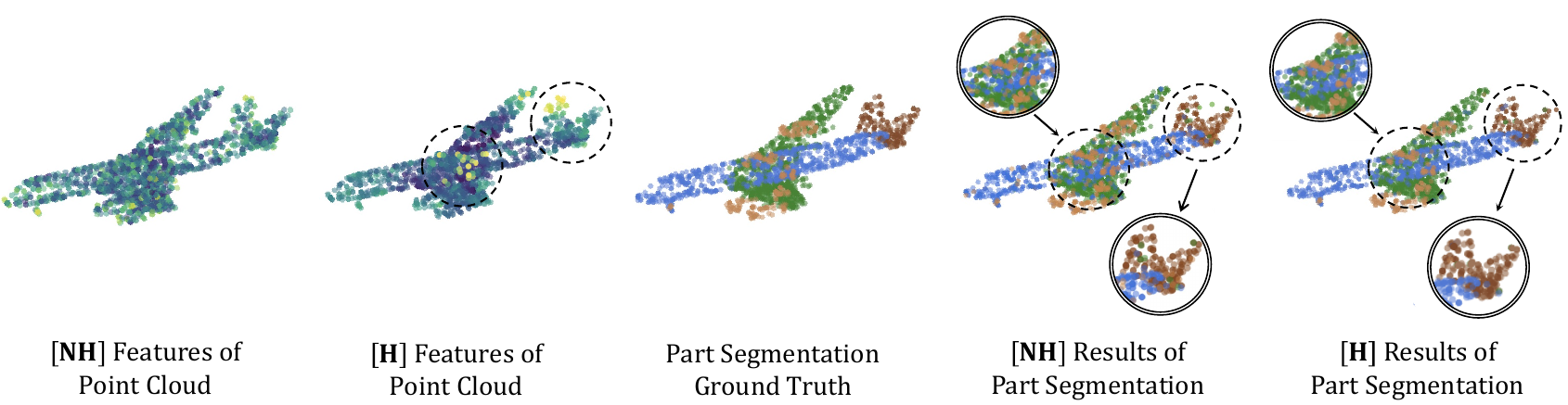}
    \vspace{-0.4cm}
  \figcaption{\textbf{Visualization of part segmentation results.} We denote the outputs from hierarchical and non-hierarchical architectures as \textbf{[NH]} and \textbf{[H]}, respectively.
   For an input point cloud (Middle), we visualize its extracted features (Left) and part segmentation results (Right).}
    \label{f7}
\end{figure}

\paragraph{Part Segmentation Results.}
The fine-grained 3D patterns learned by our hierarchical architecture largely benefits 3D downstream tasks with dense prediction, e.g., part segmentation. In Figure~\ref{f7}, we compare our Point-M2AE with multi-stage, [H], and single-scale, [NH], architectures by visualizing the extracted point features and the segmentation results on ShapeNetPart~\cite{shapenetpart}. As shown, the multi-scale architecture predicts more fine-grained part labels for the objects.




\subsection{Additional Ablation Study}
\paragraph{Transformer Stages.}
Each stage in Point-M2AE encodes the corresponding scale of the point cloud. In Table~\ref{stages}, we explore the best stage number of both encoder and decoder for learning multi-scale point cloud features during pre-training. As reported, the 3-stage encoder with 2-stage decoder performs the best. If the decoder also has three stages as the encoder, and reconstructs the point cloud at the $1$-th scale, the performance would be adversely influenced.

\paragraph{Transformer Blocks.}
In each stage, we apply several transformer blocks to encode features of the point tokens. We experiment with different block numbers in each stage of the encoder and decoder in Table~\ref{block}. We observe that stacking five blocks per stage for encoder and only one block for decoder achieve the highest accuracy. This asymmetric architecture enforces the encoder to contain more semantic information of the point cloud, which benefits the transfer capacity of Point-M2AE.

\paragraph{Fine-tuning Settings.}
\begin{wraptable}{r}{8cm}
\centering
\small
\vspace{-0.6cm}
\tabcaption{\textbf{Fine-tuning settings.} For `max + ave. pooling', we adopt max and average pooling to obtain two global features and sum them as the input of classification head. `w/o Local Atten.' denotes vanilla global self-attention.}
\vspace{0.29cm}
\begin{adjustbox}{width=\linewidth}
	\begin{tabular}{lcc}
    \toprule
    Settings &ModelNet40 &ScanObjectNN\\
    \cmidrule(lr){1-1} \cmidrule(lr){2-2} \cmidrule(lr){3-3}
     max pooling &93.3 &85.98\vspace{0.05cm}\\
     average pooling &92.8 &85.66\vspace{0.05cm}\\
     \rowcolor{gray!8}max + ave. pooling &\textbf{94.0} &\textbf{86.43}\vspace{0.05cm}\\
    class token &93.4 &86.02\vspace{0.05cm}\\
    w/o Local Atten. &93.5 &85.82\\
    \bottomrule
\end{tabular}
\end{adjustbox}
\label{finetune}
\end{wraptable}
For fine-tuning on downstream classification tasks, 
we obtain the global feature from point tokens by pooling, and apply a MLP-based head for classification. In Table~\ref{finetune}, we investigate different pooling operations along with the class token method to integrate features of all point tokens. Referring to \cite{vit}, we concatenate a learnable class token with the point tokens at the $1$-st scale, and feed them into the hierarchical encoder. After encoding, we directly utilize this class token as the global feature for classification. As reported, `max + ave. pooling' performs the best for fine-tuning, which is our default in all shape classification experiments. We also show the classification results without local spatial attention layers, which illustrates the significance of encoding local features with increasing receptive fields.

\begin{figure*}[t!]
\begin{minipage}[t!]{0.48\linewidth}
\centering
\small
\tabcaption{\textbf{Transformer stages.}
We experiment different stage number of the hierarchical encoder and decoder in Point-M2AE.}
\vspace{0.3cm}
\label{stages}
\begin{adjustbox}{width=0.8\linewidth}
\begin{tabular}{ccc}
\toprule
\ \ Encoder\ \  & \ \ Decoder\ \  &\ \ Acc. (\%)\ \ \\
\cmidrule(lr){1-2} \cmidrule(lr){3-3}
 \rowcolor{gray!8} 3 &2 &\textbf{92.9}\vspace{0.05cm}\\
2 &1 &91.8\vspace{0.05cm}\\
4 &3 &90.4\vspace{0.05cm}\\
3 &3 &90.7\vspace{0.05cm}\\
\bottomrule
\end{tabular}
\end{adjustbox}
\end{minipage}\quad
\begin{minipage}[t!]{0.48\linewidth}
\centering
\small
\tabcaption{\textbf{Transformer blocks.} 
Based on the 3-stage encoder and 2-stage decoder, we experiment different block numbers per stage.}
\vspace{0.3cm}
\label{block}
\begin{adjustbox}{width=0.8\linewidth}
\begin{tabular}{ccc}
\toprule
\ \ Encoder\ \  &\ \ Decoder\ \  &\ \ Acc. (\%)\ \ \\
\cmidrule(lr){1-2} \cmidrule(lr){3-3}
 \rowcolor{gray!8} 5 &1 &\textbf{92.9}\vspace{0.05cm}\\
4 &1 &92.7\vspace{0.05cm}\\
3 &1 &92.6\vspace{0.05cm}\\
5 &2 &91.7\vspace{0.05cm}\\
\bottomrule
\end{tabular}
\end{adjustbox}
\end{minipage}
\vspace{-0.2cm}
\end{figure*}

\paragraph{Pre-training Loss Functions.}
\begin{wraptable}{r}{7.7cm}
\centering
\small
\vspace{-0.55cm}
\tabcaption{\textbf{Pre-training losses.} `CD' and `EMD' denote Chamfer Distance and Earth Mover’s Distance losses.}
\vspace{0.3cm}
\begin{adjustbox}{width=\linewidth}
	\begin{tabular}{c c c c}
	\toprule
	 L2-norm CD &L1-norm CD &EMD &Acc. (\%)\\
	\cmidrule(lr){1-3} \cmidrule(lr){4-4}
		\rowcolor{gray!8}\checkmark &- &-  &\textbf{92.9}\\
		- &\checkmark &-  &91.1\\
		- &- &\checkmark  &91.9\\
		\checkmark &- &\checkmark  &92.4\\
		- &\checkmark &\checkmark  &91.3\\
	\bottomrule
	\end{tabular}
\end{adjustbox}
\label{loss}
\vspace{-0.2cm}
\end{wraptable}
Except for the Chamfer Distance loss~\cite{chamfer} with L2 normalization (L2-norm CD), we further evaluate the L1-normalized Chamfer Distance loss (L1-norm CD), Earth Mover's Distance loss (EMD), and their combinations. As shown in the table~\ref{loss}, the original L2-norm CD loss performs better than all other compared losses. We denote the reconstructed and ground-truth point sets as $S_1$ and $S_2$.
Compared to EMD loss that requires an optimal mapping for every point between $S_1$ and $S_2$, L2-norm CD loss only optimizes the separate pair-wise distances and is thus more robust to the variation of 3D structures. Compared to L1-norm CD loss, L2 norm of Euclidean Distances can better depict spatial distribution and pay more attention to the far away points.

\end{document}